\documentclass[a4paper]{article}

\usepackage{amssymb}
\usepackage{amsmath} % for \boldsymbol macro
\usepackage{bm}      % for \bm macro
\usepackage{graphicx}
\usepackage{float}
\usepackage{amsfonts}
\usepackage{hyperref}
\usepackage{lineno}
\usepackage{color,soul}
\usepackage[affil-it]{authblk}
\usepackage[english]{babel}
\usepackage{blindtext}

\providecommand{\keywords}[1]
{
	\small	
	\textbf{\textit{Keywords---}} #1
}

\title{Content Augmented Graph Neural Networks}

\date{}

\usepackage{amsmath}
\usepackage{bm}      % for macro
\usepackage{placeins}

\usepackage{color}
\usepackage{tabularray}

\author[1]{Fatemeh Gholamzadeh Nasrabadi\thanks{f.gholamzadeh@aut.ac.ir}}
\author[1]{AmirHossein Kashani\thanks{amkkashani@aut.ac.ir}}
\author[1]{Pegah Zahedi\thanks{pegahzahedi97@gmail.com}}
\author[1]{Mostafa Haghir Chehreghani\thanks{mostafa.chehreghani@aut.ac.ir (corresponding author)}}

\affil[1]{\small Department of Computer Engineering, Amirkabir University of Technology (Tehran Polytechnic), Tehran, Iran}

\begin{document}
	
	\maketitle
	
\begin{abstract}
 In recent years, graph neural networks (GNNs) have become a popular tool for solving various problems over graphs.
 In these models, the link structure of the graph is typically exploited and nodes' embeddings are iteratively updated based on adjacent nodes.
 %their embeddings and the embeddings of their neighbors in the previous iteration (layer).
 Nodes' contents are used solely in the form of feature vectors, served as nodes' first-layer embeddings.
 However, the filters or convolutions, applied during iterations/layers to these initial embeddings lead to their impact diminish and contribute insignificantly to the final embeddings.
 In order to address this issue, in this paper we propose augmenting nodes' embeddings by embeddings generated from their content, at higher GNN layers.
 More precisely, 
 we propose models wherein a {\em structural} embedding using a GNN and a {\em content} embedding are computed for each node.
 These two are combined using a combination layer to form the embedding of a node at a given layer layer. 
 We suggest methods such as using an auto-encoder or building a content graph, to generate {\em content} embeddings.
 In the end, by conducting experiments over several real-world datasets, we demonstrate the high accuracy and performance of our models.
 \newline
 
\end{abstract}


%\begin{CCSXML}
%	<ccs2012>
%	<concept>
%	<concept_id>10010147.10010178</concept_id>
%	<concept_desc>Computing methodologies~Artificial intelligence</concept_desc>
%	<concept_significance>500</concept_significance>
%	</concept>
%	<concept>
%	<concept_id>10010147.10010257.10010293.10010294</concept_id>
%	<concept_desc>Computing methodologies~Neural networks</concept_desc>
%	<concept_significance>500</concept_significance>
%	</concept>
%	<concept>
%	<concept_id>10002950.10003624.10003633.10010917</concept_id>
%	<concept_desc>Mathematics of computing~Graph algorithms</concept_desc>
%	<concept_significance>500</concept_significance>
%	</concept>
%	</ccs2012>
%\end{CCSXML}
%
%\ccsdesc[500]{Computing methodologies~Artificial intelligence}
%\ccsdesc[500]{Computing methodologies~Neural networks}
%\ccsdesc[500]{Mathematics of computing~Graph algorithms}



\maketitle

\section{Introduction}

Graphs are an important tool to model data and their relationships.
They are used in many domains and applications such as social networks, scientific networks, and protein protein interaction networks.
Several data analysis and machine learning tasks and problems,
including classification, regression, clustering and link prediction,
have been studied for nodes of a graph or for a collection of graphs.
% Link prediction is to predict whether two nodes in a network are like to have a link .
% It has many applications in different networks. For instance, friend recommendation, movie recommendation, knowledge graph completion, and checking interaction on protein-protein networks \cite{Zhang-18}.
% Graph data has an important role in machine learning tasks in graphs including node or graph classification, link prediction, and node clustering.
% Link prediction is one of the important tasks in network science.
%In order to study different machine learning and data analytic tasks
%(classifications, regression, clustering, link prediction, recommendation etc)
%for nodes of a graph or for a collection of graphs,
The state of the art recent algorithms for solving these problems rely on computing embeddings (representations) for nodes or graphs.
% has become popular.
In the embedding (representation) learning task, each node or each graph is mapped to a low-dimensional vector space,
so that nodes or graphs that are similar to each other in the graph space,
should find similar embeddings in the vector space.
Graph neural networks (GNNs) provide a powerful and popular tool to compute embeddings.

%There exist many graph neural networks in the literature that try to generate high quality embeddings for solving different tasks with a better performance.
A variety of graph neural networks have been proposed in the literature to generate high-quality embeddings that can enhance the performance of different tasks.
They mostly rely on the link structure of the graph and iteratively update the embedding of a node according to its own embedding and the embeddings of its neighbors in the previous iteration (layer).
As a result, the local neighborhood of a node plays a critical role in its computed final embedding.
However, nodes of a graph usually contain  a rich content that 
can be used to improve several tasks such as node classification and  clustering.
Existing GNNs mostly use this content only in the form of feature vectors
that are fed into them, as nodes' first-layer embeddings.
However, consecutive filters (convolutions) that are applied during various iterations/layers, reduce the impact of nodes' feature vectors,
so that they find little influence on nodes' final embeddings. 
As a result, the discriminative power of nodes' content information,
which can be useful in many applications, is mostly ignored.

Motivated by this observation and in order to
preserve the impact of nodes' contents at higher GNN layers,
in this paper we propose novel methods that augment nodes' embeddings with {\em nodes' content information}, at higher GNN layers.
%avoid
%diminishing the impact of nodes' contents,
%in this paper we propose augmenting node embeddings
%by using embedodings generated from their content during different GNN layers.
More precisely, 
we propose a model wherein a {\em structural} embedding using a GNN and a {\em content} embedding are computed for each node.
These two are combined using a combination layer to form the final embedding of a node at a given layer. 
We propose two methods to generate {\em content} embeddings of nodes.
In the first method, we build it
using an auto-encoder applied to nodes' initial feature vectors to improve them.
In the second method, we construct it by forming a content-similarity graph and applying a GNN to this content graph to obtain nodes' {\em content} embeddings.
%\begin{itemize}
% \item first we enhance the performance of GNNs by incorporating known techniques such as dropout and normalization, which are confirmed by our experiments as well.
% %  we modify GNNs by applying tricks that are known to be useful in improving their performance such as dropout, which is confirmed by our experiments as well.
% \item then we propose a model wherein a {\em structural} embedding using a GNN and a {\em content} embedding are computed for each node.
% These two are combined using a combination layer to form the embedding of a node at a given layer. 
% We propose two methods to generate the {\em content} embedding of nodes at a given layer.
% In the first method, we build it by improving node's initial feature vector using an auto-encoder.
% In the second method, we construct a content graph and apply GNNs on it to obtain nodes' {\em content} embeddings.
%% for each node, we extract an embedding from its content and concatenate it with the embedding generated from graph neural network.
%% We use an unsupervised dimension reduction method based on auto-encoder, to reduce the dimensions of the generated method to a desired value. 
%\end{itemize}
Our content augmentation methods are independent of the used GNN model and can be aligned with any of them.
In this paper, we apply them to three well-known graph neural networks: GCN~\cite{DBLP:conf/iclr/KipfW17}, GAT~\cite{DBLP:conf/iclr/VelickovicCCRLB18}
and GATv2~\cite{brody2022how}.
Through experiments on several real-world datasets, we demonstrate that our content augmentation techniques considerably improve the performance of GNNs. Furthermore, they outperform several state-of-the-art graph augmentation methods, such as
%By conducting experiments over several real-world datasets, we show that our content augmentation techniques considerably improve the performance of GNNs.
%Moreover, they outperform several state-of-the-art graph augmentation methods such as
LinkX \cite{lim2021large} and skip-connection \cite{you2020design}.

The rest of this paper is organized as follows.
In Section~\ref{sec:relatedwork},
we provide an overview on related work.
In Section~\ref{sec:preliminaries},
we present preliminaries and definitions used in the paper.
In Section~\ref{sec:method}, we present our methods for augmenting graph neural networks with nodes' content information.
We empirically evaluate the performance of our proposed methods in
Section~\ref{sec:exp}.
Finally, the paper is concluded in Section~\ref{sec:conclusion}.

\section{Related work}
\label{sec:relatedwork}
 
%The landscape of Graph Neural Networks (GNNs) has rapidly evolved, with significant advancements in their ability to process and interpret graph-structured data.
In this section, we provide an overview of foundational GNN models and recent innovations aimed at enhancing their performance, particularly in the context of preserving and augmenting node features.
We discuss seminal models such as Graph Convolutional Networks (GCNs), Graph Attention Networks (GATs), and their improved variant GATv2. Additionally, we explore methods to mitigate common challenges in GNNs, such as over-smoothing, through techniques like skip-connections and the LinkX model. Finally, we highlight the integration of structural and content-based features in GNNs, demonstrating their application in diverse domains.
%These discussions provide a comprehensive background for understanding the enhancements introduced by our proposed method.

%\label{sec:relatedwork2}

\subsection{GNN models}

Graph neural networks (GNNs) are designed to handle graph-structured data by iteratively updating node representations based on their neighbors. The fundamental idea is to aggregate information from a node's local neighborhood to compute its embedding.
%Our model enhances existing GNNs by preserving and augmenting nodes' content information at higher GNN layers, thus maintaining the discriminative power of nodes' features which is often diminished in traditional GNNs.
In this section, we discuss widely-used GNN models.
% that are recently proposed.
% employed in our study: GCN, GAT, and GATv2.

%\textbf{Graph Convolutional Network (GCN)}
Kipf and Welling~\cite{DBLP:conf/iclr/KipfW17} introduced graph convolutional networks (GCNs), utilizing a spectral-based approach to define convolution operations on graphs. The message-passing mechanism in GCNs can be described as follows: node features are aggregated from neighboring nodes using a normalized adjacency matrix with added self-loops. This normalization aids in maintaining numerical stability and enhances the propagation of information through the network layers. The propagation rule can be defined as:
\begin{equation*}
	H^{(l+1)} = \sigma\left(\tilde{D}^{-1/2}\tilde{A}\tilde{D}^{-1/2}H^{(l)}W^{(l)}\right),
\end{equation*}
where $\tilde{A} = A + I$ is the adjacency matrix with added self-loops,
$\tilde{D}$ is the degree matrix of $\tilde{A}$, $H^{(l)}$ is the feature matrix at layer $l$, $W^{(l)}$ is the layer-specific trainable weight matrix, and $\sigma$ is an activation function.
%\textbf{Graph Attention Network (GAT)}

Velickovic~et~al.~\cite{DBLP:conf/iclr/VelickovicCCRLB18} introduced graph attention networks (GATs), which employ an attention mechanism to assign varying weights to different nodes in the neighborhood. This attention mechanism enables the model to focus on the most relevant parts of the graph. In GATs, attention coefficients are calculated for each edge, and these coefficients are used to weigh the node features from neighboring nodes during aggregation.
The attention coefficient $\alpha^{(l)}_{vu}$ between nodes $v$ and $u$ 
at layer $l$
is defined as:
\begin{equation*}
	e^{(l)}_{vu} = \text{LeakyReLU}\left(a^T \left[W h^{(l)}_v || W h^{(l)}_u \right]\right),
\end{equation*}
where $a$ is a learnable weight vector, $W$ is a shared weight matrix, $||$ denotes concatenation, and $h^{(l)}_v$, $h^{(l)}_u$ are the feature vectors of nodes $v$ and $u$ at layer $l$.
The normalized attention coefficients $\alpha^{(l)}_{vu}$ are then computed using a softmax function:
\begin{equation*}
	\alpha^{(l)}_{vu} = \frac{\exp\left(e^{(l)}_{vu}\right)}{\sum_{k \in \mathcal{N}(v)} \exp\left(e^{(l)}_{vk}\right)}.
\end{equation*}
The node feature update is then given by:
\begin{equation*}
	h^{(l+1)}_v = \sigma\left( \sum_{u \in \mathcal{N}(v)} \alpha^{(l)}_{vu} W h^{(l)}_u \right).
\end{equation*}
%\textbf{GATv2}
Brody~et~al.~\cite{brody2022how} proposed GATv2, an extension of GAT, which addresses limitations in the original GAT by introducing more expressive attention mechanisms. GATv2 enhances the attention computation process, providing a more flexible and powerful way to aggregate node features based on their importance as determined by the attention mechanism. The attention mechanism in GATv2 is defined as:
\begin{equation*}
	e^{(l)}_{vu} = a^T\text{LeakyReLU}\left(W \left[h^{(l)}_v ||  h^{(l)}_u \right]\right),
\end{equation*}
where $a$ is a learnable weight vector and $W$ is a shared weight matrix.
%The rest of the attention computation and node feature update follows the same process as in GAT, with the improved attention mechanism allowing for more complex relationship capture.

%\subsection{GNNs with content-based features}

\subsection{Mitigating over-smoothing in GNNs}
%\hfill \break

Over-smoothing is a common issue in deep GNNs,
wherein node representations become indistinguishable as the number of layers increases. 
Alon and Yahav~\cite{alon2020bottleneck} studied the challenge of over-squashing in deep graph neural networks, a problem that intensifies with the growing number of GNN layers.
They suggested maintaining essential information flow from the initial layers to a central bottleneck layer. Additionally, they proposed a mechanism to create new edges, connecting crucial nodes to the target node.
%This innovative approach relocated pivotal nodes to higher layers, enhancing the network's capability to absorb and effectively utilize crucial data during the propagation process.
%\textbf{Skip-Connections}
investigated the use of skip-connections as a simple yet powerful method to address over-smoothing. By creating direct links between non-adjacent layers, skip-connections allow information to bypass certain layers, thus maintaining node-specific features that could be lost in deeper networks.
This approach tries to preserve the uniqueness of node representations by ensuring that the initial features are incorporated into the final embeddings.
The propagation rule with skip-connections can be described as:
\begin{equation*}
	H^{(l+1)} = \sigma\left(\tilde{D}^{-1/2}\tilde{A}\tilde{D}^{-1/2}H^{(l)}W^{(l)}\right) + H^{(l)},
\end{equation*}
where the added $H^{(l)}$ term represents the skip-connection from layer $l$ to layer $l+1$.

%\textbf{LINKX}
One of the works most similar to ours is LinkX.
Lim et al.~\cite{lim2021large} aimed to address the limitations of existing GNNs in non-homophilous graphs by introducing the LinkX method. LinkX separately embeds adjacency and node features and combines them using multilayer perceptrons (MLPs).
%One of most similar works to our work is LinkX.
%Lim et al.~\cite{lim2021large} tried to address the limitations of existing GNNs in non-homophilous graphs by introducing
%%a collection of large, diverse datasets and
%the LinkX method. LinkX embeds adjacency and node features separately and combines them with multilayer perceptrons (MLPs).
It employs the following feature update mechanism:
\begin{equation*}
	H^{(l+1)} = \sigma\left(\tilde{D}^{-1/2}\tilde{A}\tilde{D}^{-1/2}H^{(l)}W^{(l)}\right) + XW^{(0)},
\end{equation*}
where $X$ is the original feature matrix and $W^{(0)}$ is a trainable weight matrix applied to input features.
%Compared to skip-connections, our proposed method transmits information in a compact and optimal way to the next layers. This approach prevents the model from becoming biased based on redundant information, thereby enhancing the decision-making process by focusing on the most relevant data.
Our approach diverges from LinkX by not only mitigating over-smoothing but also enhancing the discriminative power of node features through content augmentation. While LinkX focuses primarily on preserving the initial node features, our models integrate additional content-based features at higher GNN layers, thereby enriching the node representations with more contextual information. This dual focus on preserving and augmenting node features allows our models to maintain high discriminative power even in deeper GNN architectures.

Recently, several other techniques such as strong transitivity relations \cite{DBLP:journals/corr/abs-2401-01384},
balanced Forman curvature~\cite{10.1007/978-3-031-53468-3_19}, 
positional encodings \cite{positional},
discrete geometry \cite{10.1007/978-3-031-53468-3_19} and
central nodes \cite{Zohrabi2024} have been employed to augment graph structures and improve the performance of GNNs.
Our {\em content} augmentation methods can be easily aligned with these
{\em structural} augmentation techniques to enhance their effectiveness.

\subsection{Fusing structural and content-based features}
%\hfill \break
In recent years, the integration of structural and textual data
for different tasks over graphs
%for link prediction in online social networks
has gained significant attention. Dileo et al.~\cite{dileo2024temporal} proposed a temporal graph learning approach that leverages both graph structure and user-generated textual content for dynamic link prediction in online social networks. Their methodology incorporates BERT language models to process textual data and combines it with dynamic GNNs to predict future links. This work is particularly relevant to ours as it highlights the importance of nodes' textual information in predicting link formation.

In recommendation systems, hybrid models that combine GNNs with other representation techniques have shown promise. Spillo et al.~\cite{spillo2023combining} introduced a knowledge-aware recommender system (KARS) that integrates GNNs and sentence encoders. Their approach first uses GNNs to learn embeddings from collaborative filtering data and descriptive features. Then, a sentence encoder processes textual content to learn representations. These embeddings are further refined using self-attention and cross-attention mechanisms to predict user preferences.
Jin et~al.~\cite{DBLP:conf/wsdm/Jin211} introduced SimP-GCN, a graph convolutional network with the following steps. First, it uses an adaptive technique to fuse structural and node features. Next, it employs a learning approach to predict pairwise feature similarity based on the hidden embeddings of selected node pairs.
%Through this approach, SimP-GCN endeavors to uphold both feature and structural resemblances while constructing node embeddings.

In the GIANT model~\cite{chien2021node}, the authors tackled feature scarcity in graph-agnostic contexts by using a pre-trained BERT model to predict neighborhoods from raw textual data. They extracted feature vectors from the BERT model and seamlessly integrated them into the GNN model.
%, effectively enhancing the model's capability to bridge between text and graph domains.
Sawhney et al.~\cite{sawhney2020deep} proposed a model that integrates deep learning techniques with attention mechanisms to predict stock movements. This model combines financial data, social media text, and inter-company relationships to create a comprehensive prediction tool. It emphasizes the importance of multimodal data and the use of attention mechanisms to capture nuances in stock movement prediction.

\section{Preliminaries}
\label{sec:preliminaries}

%Throughout the paper,
%we use the following standard for notations and symbols:
%lowercase letters for scalars,
%uppercase letters for sets, multisets and graphs,
%bold lowercase letters for vectors
%and bold uppercase letters for matrices.
We assume that the reader is familiar with basic concepts in graph theory.
% In this section, we introduce the definition of network embedding, link prediction, and then provide definition different centrality measures that in our method used.
% \subsection{Notation}
% In this section, we introduce the definition of network embedding and link prediction and then provide definition different centrality measure that in our method used.
By $G =(V,E)$, we refer to a graph whose node set is $V$
and edge set is $E$.
%We use $n$ to denote the number of nodes of $G$.
%We may represent $G$ by an adjacency matrix $A \in \mathbb R^{n \times n}$,
%where $A_{ij}$ indicates whether there is an edge between nodes $i$ and $j$.
% are connected represents the similarity of $v_i$ and $v_j$ on the graph.
%We use $d_{uv}$ to refer to the distance (shortest path length)
%between nodes $u$ and $v$.
%Furthermore, we denote the degree and out degree of node $v$
%by $deg(v)$ and $deg^{out}(v)$, respectively.
By $n$ and $m$, we respectively denote the number of nodes and the number of edges of the graph.
By $n_t$ and $m_t$ we respectively denote the number of
training nodes and the number of training edges of the graph.
We assume that each node has a feature vector of size $d$.
%and d is the dimensions of the node feature vectors.
By $X \in \mathbb R^{n \times d}$, we denote the matrix of nodes'
feature vectors.
By finding vector embeddings for nodes of a graph,
we want to encode them as low-dimensional vectors that summarize the structure of the graph.
% Node embedding takes an input graph and a parameter $d$ as the dimensionality of the node embedding.
A node embedding function $g$ is a mapping
%$g:V \rightarrow \mathbb R^d$
that maps each node in the graph
into a low-dimensional vector of real values \cite{node2vec-16}.
The most popular methods to generate vector embeddings are graph neural networks.
%We denote the embedding vector of a node $u$
%with ${f}(u)$.
%and its $i$-th entry with $f_i(u)$.
%With $N_s(v)$,
%we refer to the neighbourhood of node $v$,
%obtained by sampling strategy $s$.
% $exp()$ indicates softmax unit.
% We also use $d_{tx}$ and $\alpha(t,x)$ for show distance and unnormalized transition probability from node $t$ to node $x$, respectively.

At a high level, GNNs consist of the
following steps~\cite{Chehreghani_nmi}:
\begin{enumerate}
	\item[i)]
	for each node $v$, a neighborhood $N(v)$ is constructed,
	\item [ii)] at the first layer, the embedding of node $v$ consists of its features (or e.g., the zero vector, if attributes are absent),
	and
	\item[iii)] at each layer $l+1$, the embedding of node $v$ is computed
	using a function $ f$ that takes as input the layer-$l$ embedding of $v$
	and the layer-$l$ embeddings of the nodes in the
	neighborhood of $v$.
	Function $f$ consists of the following components:
	i) a linear transformation defined by (at least) one trainable weight matrix $W$ that converts a lower layer embedding into a {\em message},
	ii) an activation function $\sigma$, which is element-wise  applied to each generated message to induce non-linearity to the model, and
	iii) an aggregation function $agg$ which takes a number of vectors as input, aggregates them, and generates a vector as the higher layer embedding of the node.
\end{enumerate}
As a result, function $f$ is defined as follows:
\begin{equation}
	\label{eq:gnn}
	 h_v^{(l+1)} = f(v,N(v)) =
	\mathsf{agg}\left( \sigma \left(  W \cdot   h_u^{(l)} , \forall u \in N(v) \cup \{v\}
	\right)\right),
\end{equation}
where $ h_v^{(l+1)}$ and $h_v^{(l)}$ are respectively the vector embeddings of $v$ at layers $l+1$ and $l$.
A widely used activation function is ReLU, defined as follows:
${ReLU}(x) = {max} (0,x)$.

\section{Content augmented graph neural networks}
\label{sec:method}

%In this paper, we present two methods to address a challenge faced by GNNs in utilizing content information, effectively.
As already mentioned, 
our primary concern in this paper lies in the vanishing of essential first-layer information, specifically content information, as it propagates through multi-layer convolutional GNNs. Consequently, the models tend to over-rely on graph structural information during the final decision-making process
and pay less attention to nodes' content.
In this section, we present two effective models that are designed to augment GNNs with content information.
The first model is proper for the supervised setting, wherein enough labeled examples should be accessible during training.
This is because of the auto-encoder, utilized within this model: although its encoder/decoder components are inherently unsupervised, in our model,
they should be trained with enough number of examples to learn dimension reduction patterns, effectively.
%Consequently, this model achieves high performance when training data is scarce.
%, as is the case in semi-supervised methods. 
The second model works fine in the semi-supervised setting, wherein
a limited amount of labeled data is accessible during training.
% one designed for semi-supervised setting and the other for supervised setting, aimed at enhancing the performance of classification tasks.
%In contrast, the second model
It constructs a content graph, alongside the input graph, and applies GNN models to both of them. GNNs are known for their superior performance in the semi-supervised setting. As a result, the second model demonstrates outstanding performance 
in this setting.
%in the semi-supervised tasks.

\begin{figure}[htbp]
    \centering
    \includegraphics[scale=0.3]{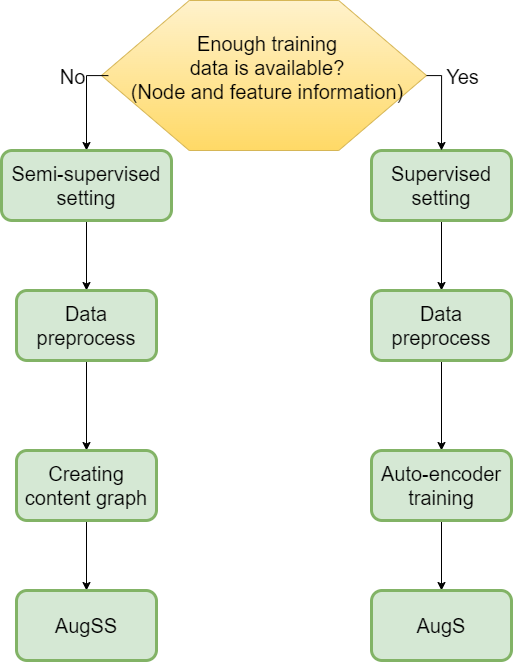}
    \caption{High-level description of content augmentation by our proposed methods.}
    \label{fig:abstract_flowChart}
\end{figure}

In the rest of this section, first, we briefly discuss how the input graph
is preprocessed, before feeding it into GNN models.
Then we describe our supervised and semi-supervised content augmentation methods. Figure~\ref{fig:abstract_flowChart} presents a high-level demonstration of content augmentation performed by our proposed models.

% \begin{figure}[H]
% 	\centering
% 	\includegraphics[width= \textwidth]{Abstract NLP paper.drawio (4).png}
% 	\caption{Abstract Flowchart on How to Use the Suggested Models in This PaperThe high-level architecture of AugS-GNN. Red units are computational and blue units are data.}
% 	\label{fig:abstract_flowChart}
% \end{figure}

%Finally, we discuss how the generated embeddings are used for a downstream task, such as node classification.

\subsection{Preprocessing}
\hfil\\
In our data preprocessing phase, we adopt a bag-of-words approach~\cite{le2014distributed} for content processing, creating the initial vectors that will be employed by the GNN layers. The bag-of-words technique represents the content information of the input data, serving as a foundation for subsequent graph-based operations. It should be noted that due to variations in the datasets, this approach results in different first-layer vector sizes. We will elaborate on this in Section~\ref{sec:exp}, providing insights into the impact of diverse dataset characteristics on GNNs' performance.
%We normalize 

Furthermore, to enhance information flow and preserve local content within the graph, it is common practice to augment the graph structure by adding self-loops to each node. Our experiments demonstrate that the addition of self-loops generally improves performance, particularly in the supervised setting where sufficient training data is available. However, an exception to this is the Cora dataset, which features a low average degree and the smallest number of nodes and features. This exception will be discussed further in the experiments section. Adding self-loops does not necessarily lead to performance improvement in the semi-supervised setting
(therefore, in the experiments section, we do not report on additional self-loop experiments in the semi-supervised setting).
%(therefore, we do not report on additional self-connection experiments in semi-supervised contexts, as they typically show worse performance).
%setups is incorrect due to a lack of sufficient data points for effective training.

\subsection{Supervised content augmentation of graph neural networks (AugS)}

\begin{figure}[H]
	\centering
	\includegraphics[width= \textwidth]{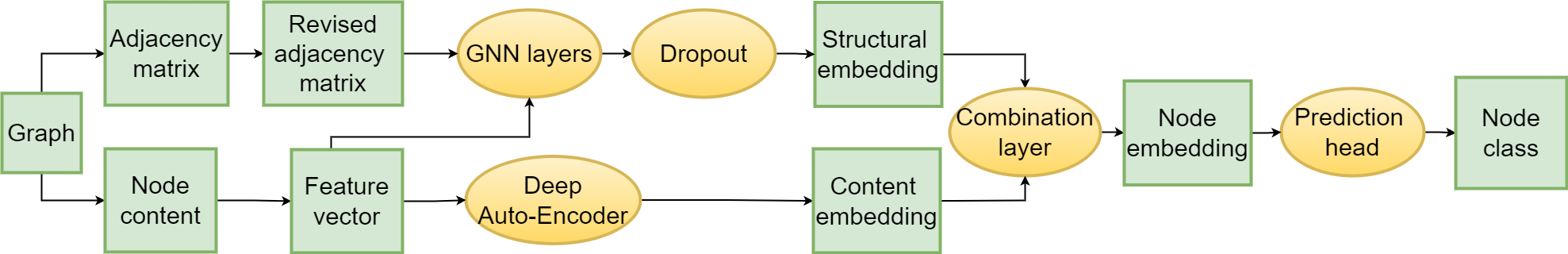}
	\caption{High-level structure of AugS-GNN. Yellow units are computational and green units are data.}
	\label{fig:enter-label}
\end{figure}

%In the supervised setting we propose a novel architecture that ensures the proper integration of content information throughout the GNN layers, thereby facilitating better utilization of content data.
Figure~\ref{fig:enter-label} presents the high level structure of our proposed  model
for content augmentation of graph neural networks, in the supervised setting.
This model is compatible with any graph neural network and does not depend on the specific model used. As mentioned earlier, this model is primarily effective in supervised scenarios, as deep auto-encoders tend to have limited functionality in cases of data scarcity.
%our content augmentation  method is independent from the used graph neural network
%and can be adapted with any of them.
We refer to this supervised content augmentation model as AugS-GNN.
In the following, we describe each component of the model in details\footnote{Our implementation of AugS is publicly available at: \url{https://github.com/amkkashani/AugS-GNN}}.

\subsubsection{Structural and content embeddings}
\hfil\\
As mentioned earlier, we create two distinct embeddings for each node.
First, we utilize the GNN model, which incorporates both the graph structure and features generated from the bag-of-words approach.
%In this paper, we refer to this model as GNN* (or GCN*, when the used GNN model is GCN).
In this way, a {\em structural} embedding is built for each node.
Then, in order to add a stronger dimension of content information,
at higher GNN layers and for each node,
we generate a {\em content} embedding.
This is done by feeding the first-layer embedding (the initial feature vector) 
of each node into an auto-encoder.
The output of the encoder component of the auto-encoder
serves as the {\em content} embedding.
%word embeddings  created using an encoder, adding another dimension of content information to our model.
%The combination of these embeddings from both GNN and the encoder results in a powerful representation of the data.
%In this way, our model can effectively leverage both graph structure and content information during the embedding learning process.
%contributing to its superior performance and adaptability across various tasks.

For auto-encoder,
we use the model based on multiple MLP encoder layers and
multiple MLP decoder layers, introduced by Hinton~\cite{hinton2006reducing}.
In the encoder part, each layer's size is reduced by half compared to the previous layer, while the decoder part follows an opposite pattern, with each layer's size being doubled relative to the preceding one. The unsupervised loss function used during training is defined by Equation~\ref{eq:loss_enc_dec_function},
which helps train the model's parameters:
\begin{equation}
	\label{eq:loss_enc_dec_function}
	J(\theta) = \sum_{i = 0}^{n_t} ||f_{dec} \circ f_{enc}(x_{i}) - x_{i}  ||_{2}^{2},
\end{equation}
where $\Theta$ is the set of trainable parameters of the encoders and decoders, operator $\circ$ is function composition,
$f_{\text{enc}}$ and $f_{\text{dec}}$ respectively denote the encoder and decoder functions (each consists of several MLP layers) and
$x_{i}$ represents the initial feature vector of node $i$.
The loss function computes the squared L2 norm of the distance vector between the
decoder's composition of the encoder's output and the original input.
Using it, we try to train the auto-encoder parameters in a way that each vector, 
after encoding and decoding, becomes close to its original form as much as possible.
%make the measures how close the model's output is to the input and provides feedback for training the model.
The summation over $n_t$ considers feature vectors of all nodes of the training dataset. 
After completing the training phase, we use the output of the last encoder (the input of the first decoder), as the {\em content} embedding.
This layer is called the {\em bottleneck} layer.
It is worth highlighting that the parameters of the auto-encoder are not jointly learned with the parameters of the entire model, as a separate loss function is used to train them.
We set the input dimension based on the number of node features and the bottleneck dimension to $64$.

%We use the following setting to train the auto-encoder used to generate {\em content} embeddings.
%We set the number of epochs to $1000$,
%the input dimension based on number of features and
%the bottleneck dimension to $64$.
%%We use the adadelta algorithm~\cite{} as the optimizer, 
%We use Adam as the optimizer,
%ReLU as the activation function in the hidden layers and softmax as the activation function in the output layer.

%\begin{table}[h]
%\caption {Encoder} \label{tab:encode} 
%\begin{tabular}{|cc|}
%\hline
%Epoch                 & 50                    \\ \hline
%Batch Size            & 32                    \\ \hline
%Loss Function         & Mean square error     \\ \hline
%Optimizer             & adadelta              \\ \hline
%Activation Function   & Relu, Softmax(output) \\ \hline
%Input dimension       & 256                   \\ \hline
%Bottleneck dimenstion & 16                    \\ \hline
%\end{tabular}
%\end{table}

%extract the encoder part of the model to generate the second embedding for our model.
%where $x$ represents feature vectors generated by the bag of words approach,.

\subsubsection{Combination layer}
\hfil\\
In this layer, we combine the {\em structural} and {\em content} embeddings obtained for each node,
to form a unique embedding for it.
%The combination of the obtained embeddings stands as a crucial focal point in this research, as it greatly influences the utilization of content information. By effectively combining the embeddings from the GNN model and the encoder, we aim to enhance the model's ability to capture relevant content features.
Our combination layer consists of two phases: the fusion phase, wherein the two structural and content embeddings are fused to form a single vector;
and the dimension reduction phase, wherein the dimentionality of the vector obtained from the first phase is reduced.
For the first phase, we have several fusion functions, including:
{\em concatenation} where the two vectors are concatenated,
{\em sum} where an element-wise sum is applied to the vectors
and {\em max} where an element-wise max is applied to the vectors.
Our experiments demonstrate that concatenation consistently outperforms the other fusion methods across different datasets.
Therefore in this paper, we specifically highlight the results achieved using the concatenation function.
%showcasing its superior performance in leveraging content information for improved model accuracy and effectiveness. 

For the second phase, we incorporate an MLP, whose parameters
are trained jointly with the other parameters of the model (unlike the parameters of the auto-encoder used to generate the {\em content} embedding).
% to further process the fused embeddings.

\subsubsection{Prediction head}
\hfil\\
The embeddings generated by AugS-GNN can be used as the input for several downstream tasks and problems such as classification, regression, and sequence labeling.
%serve various purposes, including classification, regression, sequence labeling, and other downstream tasks.
The method used to address these tasks is called the prediction head.
Various machine learning techniques, such as SVD, decision trees, and linear regression, can be used in the prediction process. In the experiments of this paper, our focus is to assess the model's performance in classifying nodes. To do so, 
we use an MLP, consisting of two dense hidden layers each one with 16 units, as the prediction head.
The parameters of the prediction head are jointly learned with the other parameters of the whole model. 
%
%By employing the fused embeddings from both the GNN model and the encoder, our model gains a robust representation of the input data, enriched with relevant content information. This comprehensive representation enables the model to make informed decisions, making it well-suited for classification tasks across diverse datasets. Through extensive experimentation and evaluation, we assess the classification performance of our model and demonstrate its effectiveness in handling various classification tasks with superior accuracy and generalization capabilities.

We train our model using the cross entropy loss function~\cite{shannon1948mathematical}.
For a single training example, it is defined as follows:
\begin{equation}
	-\sum_{k=1}^{c}  T_{k} \log(S_{k}),	
\end{equation}	
where $c$ is the number of classes
and \( T_k \) and  \( S_k \) respectively represent the true probability and
the estimated probability of belonging the example to class $k$.
The total cross entropy is defined as the sum of cross entropies of all training examples.

%, defined as follows~\cite{shannon1948mathematical}:
%\begin{equation}
%	D(S,L) = -\sum_{i}^{} L_{i} \log(S_{i})
%\end{equation}
%Where \( L_i \) represents the true label of the $i$-th example of the training dataset and
%\( S_i \) indicates its predicted label.

We use the following setting to train the model.
We set the number of epochs to $200$,
batch size to $32$,
dropout ratio to $0.05$ (in the MLP of the prediction head, we set it to $0.2$),
and the number of GNN layers to $2$.
We use the Adam algorithm as the optimizer and
ReLU  as the activation function in the hidden layers and softmax in the output layer.
%We use early stopping, based on the validation accuracy.
%As already mentioned, the initial feature vectors of nodes are derived from their textual content, using a bag-of-words representation that counts the occurrence of each word.

\subsubsection{Complexity analysis}
\hfil\\

%In this section, we discuss both time complexity and the number of parameters of our model.
%In our notation, \( N \) denotes the number of nodes, \( E \) represents the number of edges, and \( F \) stands for the number of features in a layer. Specifically, \( F_{\text{input}} \) or \( F_0 \) refers to the number of features in the input layer, while \( F_{\text{output}} \) indicates the number of features in the output vector.

As can be seen in Figure~\ref{fig:enter-label}, our model consists of three parts:
the GNN part, the auto-encoder part, and the prediction head.

\begin{itemize}
\item
  Time complexity of the GNN part for \( L \) layers of computation is \cite{DBLP:conf/iclr/VelickovicCCRLB18}:
$$ O\left(\sum_{l=1}^{L} \left( m_t d_{l-1} + n_t d_{l-1} d_l \right)\right), $$
where $n_t$ and $m_t$ are respectively the number of training nodes 
and the number of training edges, 
and $d_{l}$ is the size of embeddings at layer $l$.
\item
In the auto-encoder part,
due to the symmetric architecture we use,
time complexity and the number of parameters of the encoders are the same as those of the decoders.
Therefore, we focus only on the encoders.
%and multiply its complexity by 2 to calculate the complexity of the entire part. This multiplication by 2 does not affect the final upper bound of the complexity.
In the encoding part, let \(d_{\text{input}}\) and \(d_{\text{bot}}\)
be respectively the size of input layer and the bottleneck layer.
Since each layer in the encoding part halves the number of features of the previous layer,
the depth of the encoder is
$$ \left\lceil \log\left( \frac{  d_{\text{input}} }{ d_{\text{bot}} } \right)  \right\rceil + 1 .$$
%\(\log(|F_{\text{input}}|) - \log(|F_{\text{bot}}|)\).
%We use the largest power of 2 that is less than half of \(F_{\text{input}}\). We continue this process until we reach the hidden embedding size.
%Therefore, the depth of our encoder is
%\(\log(|F_{\text{input}}|) - \log(|F_{\text{hidden}}|)\) 
Since this number is small, we consider it a constant.
On the other hand, time complexity of a MLP is 
\[
O\left(\sum_{i=1}^{k} n_t d_{i-1} d_i\right),
\]
where \( k \) is the number of layers in the MLP and is a constant, \( d_{i-1} \) is the number of input features to layer \( i \), and \( d_i \) is the number of output features of layer \( i \).
Letting  ${d_{\text{max}}}$ be the maximum size that a layer may have,
time complexity of the auto-encoder becomes \( O\left( n_t {d_{\text{max}}}^2 \right) \).
%The upper bound of the time complexity is \( O(|F_{\text{max}}|^2) \), which is the maximum feature size of a layer in our neural network.
Similarly, it can be shown that the number of parameters of the auto-encoder is \( O\left({d_{\text{max}}}^2\right) \).
%Additionally, to calculate the upper bound of our model, In our models, \( F_{\text{max}} \) is equal to \( F_{\text{input}} \). However, in the general case where we do not have sufficient initial features in our input layers, this assumption does not hold true.
%For the GNN part, the complexity could be \( O(|E| |F_{\text{max}}| + |N| |F_{\text{max}}|^2) \), and for the encoder section, it could be \( O(|F_{\text{max}}|^2) \)
\item
Time complexity of the
 prediction head, which is a MLP, 
is  \( O\left(n_t {d_{\text{max}}}^2 \right) \).
\end{itemize}
Therefore, the overall time complexity of our model is:
\begin{equation}
%	O(|E| d_{\text{max}} + n_t d_{\text{max}}^2 + 2 d_{\text{max}}^2) = 
O\left(m_t {d_{\text{max}}} + n_t {d_{\text{max}}}^2 \right).
\end{equation}
As a result,
our supervised content augmentation method does not increase time complexity order of GNN models.

A single GNN layer and a single MLP layer have the same order of parameters, which is \( O\left({d_{\text{max}}}^2\right) \).
Moreover, we consider the depths of all the used neural networks constant. Therefore, the parameter complexity of the GNN, the auto-encoder, and the prediction head is \( O\left({d_{\text{max}}}^2\right) \).
Thus, the overall parameter complexity of our method is \( O\left({d_{\text{max}}}^2\right) \), which is the same as the parameter complexity of the original GNN models.

\subsection{Semi-supervised content augmentation of graph neural networks (AugSS)}

In this section, we present our second content augmentation method,
specifically tailored to work in a semi-supervised setting.
This approach involves the construction of an auxiliary graph based on nodes'
content attributes, which is subsequently integrated with the input graph during GNN processing. By adopting this strategy, we infuse the GNN with the influence of nodes' contents, effectively mitigating the issue of content information degradation within the GNNs.
We refer to this semi-supervised content augmentation model as AugSS-GNN.
%It is important to note that, 
Similar to AugS, this approach is compatible with any graph neural network and does not rely on the particular model employed\footnote{Our implementation of AugSS is publicly available at: \url{https://github.com/FatemehGholamzadeh/AugSS-GNN}}.

\subsubsection{Content graph construction}
\hfil\\
%As previously stated,
Our second approach for content augmentation in GNNs involves the creation of a "content graph", which is fed into the GNN framework. Henceforth, in this paper, we refer to the input graph as the "structural graph", to distinguish it  from the content graph. The construction of the content graph is done through the utilization of inherent features within the graph nodes.
Here are two steps for building the content graph:

\begin{enumerate}
	\item 
	We compute pairwise similarities between initial feature vectors of nodes.
	%	These similarity values encompasses
	We can employ diverse metrics suitable for assessing the similarly between  two vectors, such as Euclidean distance, dot product, or cosine similarity. In this paper, we employ cosine similarity.
	%	 due to its computational efficiency.
	
	\item 
	If the similarity value of the feature vectors of two nodes exceeds a certain threshold $\epsilon$,
	we create an edge between the two nodes.
	%	corresponding to those feature vectors.
	%	Conversely, if the similarity measure between two feature vectors falls below the specified threshold, no edge is established between the corresponding nodes.
	It is important to note that the threshold value varies across different datasets.
	In this study, we employ a grid search methodology to determine the optimal threshold for each dataset.
\end{enumerate}

\subsubsection{Model's architecture}
\begin{figure}[H]
	\centering
	\includegraphics[width= \textwidth]{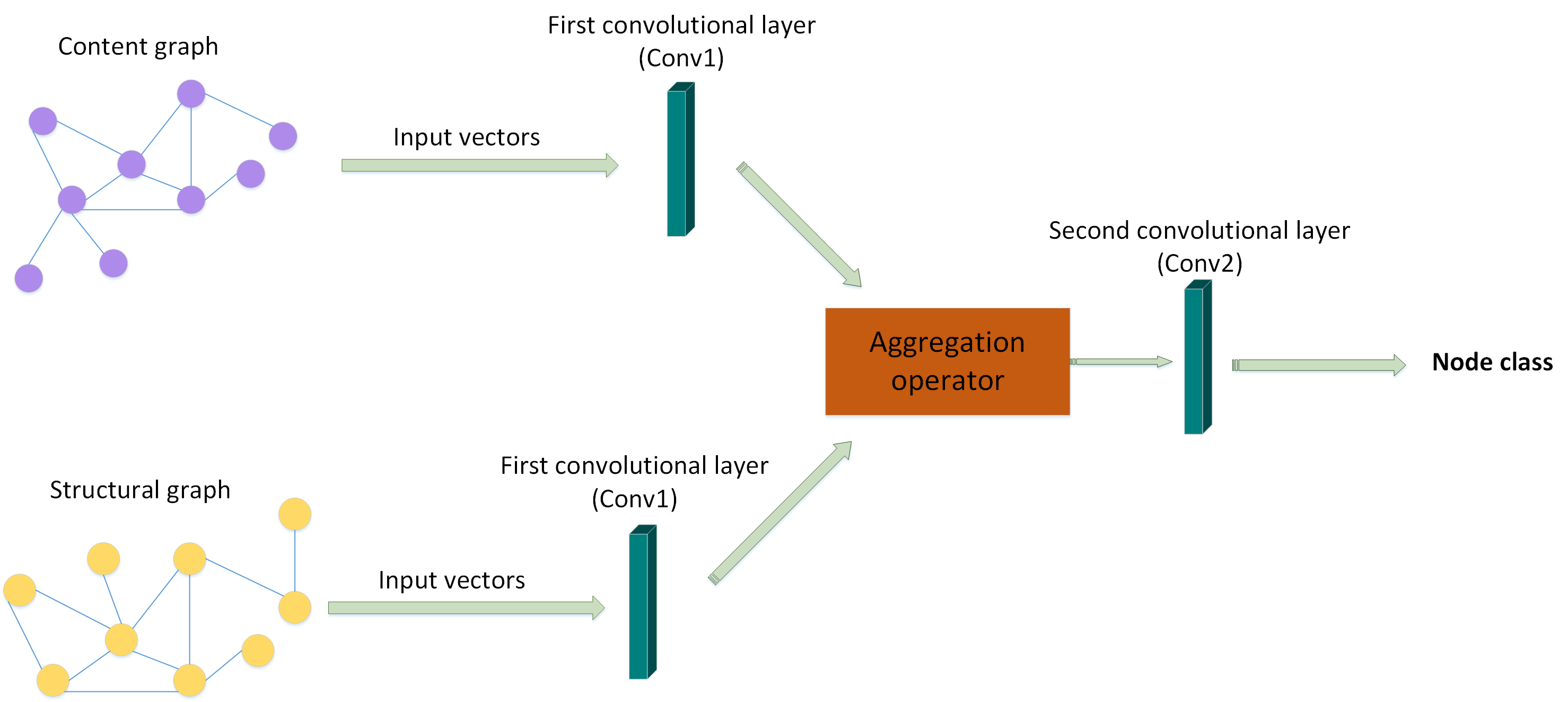}
	\caption{The high-level architecture of AugSS-GNN.\label{fig:aug_2} }
\end{figure}

\hfil\\
%In this section, we describe the architecture of the model and its utilization of the content graph.
Our semi-supervised strategy for integrating the content graph into the training procedure of graph neural networks is illustrated in Figure~\ref{fig:aug_2}.
Apart from the structural graph, we introduce the content graph as an input to the graph neural network. GNNs take a feature matrix of graph nodes as input. Therefore, both the structural and content graphs require an initial feature matrix to be provided as input to the GNN. In our approach, we employ the same feature matrix $X$ as the initial feature matrix, for both the structural graph and the content graph. We examined alternative techniques, such as vectors generated by the DeepWalk algorithm \cite{DeepWalk-14},
or initializing the feature matrix for the content graph with a set of structural graph features.
% However, these methods did not yield satisfactory results.
%Hence, we employ  matrix $X$ as the initial feature matrix of the structural graph.
We will discuss their results in Section \ref{sec:exp}.
Let's denote the structural graph by $G$ and the content graph by $G'$.
All graph neural networks used in this section have two graph convolution layers.
It is important to note that each of these graph convolution layers has its own weight parameters, and they do not share the trainable weight parameters.

 First, a graph convolution layer, called Conv1, is applied to each of the graphs, considering the adjacency matrix specific to that graph. 
%among them.
%For example, if the input graph $G$ has a feature matrix $X$, and the input graph $G'$ has a feature matrix $X'$, the matrix $X$ is fed as input to the Conv1 layer, which is the first layer of the graph neural network. The Conv1 layer is applied to $X'$ as well. However, since we decide to consider the same initial node features for both graphs, %matrices X and X' are the same in our method, meaning:
%matrices $X$ and $X'$ are identical and represent the initial node features.
%$X = X' = initial node features$
Suppose that Conv1 maps input vectors with dimension $d$ to output vectors with dimension $d'$.
After applying this layer to $G$ and $G'$, 
%he first graph convolution to matrix $X$
%(considering node adjacencies in graph $G$), and once again to matrix $X'$
%(considering node adjacencies in graph $G'$),
we will have two output matrices with the same size $n_t \times d'$.
In fact, for each node we will have two vectors of dimension $d'$.
At this stage, we combine these two vectors using an aggregation method,
which could be, for example, element-wise averaging or summing,
or their concatenation, to obtain a vector of dimension $2d'$. Then, we reduce the dimensionality of this vector to obtain a vector of dimensions $d'$, which is used as input to the next layer.

In our experiments, 
we also employed an alternative aggregation method that led to improved results. We introduced two scalar trainable parameters, $w_1$ and $w_2$, serving as the weights for the structural graph and content graph outputs, respectively. As a result, if we denote the structural graph output with $h_1$ and the content graph output with $h_2$, the combined output can be computed as follows:
\begin{equation}
	h = w_1 \times h_1 + w_2 \times h_2 .
\end{equation}
We train these two weights throughout the network's training process,
along with the other parameters.
This dynamic training approach allows us to adaptively determine the significance of each embedding component.

The second layer of the graph neural network (Conv2), which is also the final layer, takes input vectors of dimension $d'$ and maps them to vectors of dimension $c$, where $c$ is the number of classes in node classification tasks.
Therefore, by applying the softmax function, the class of each node is determined.
Similar to the AugS method, we use cross entropy as the loss function.
Figure \ref{fig:aug_2} illustrates our second model's architecture.

\subsubsection{Complexity analysis}
\hfil\\

AugSS employs two distinct graph convolution layers, applied to both the structural graph \( G \) and the content graph \( G' \).
%, each with its own weight parameters. 

\begin{itemize}
\item
The first convolution layer transforms input vectors of dimension \( {d }\) to output vectors of dimension
\( d' \). %Considering that the upper bound for input features is \( |F_{\text{max}}| \),
Therefore, time complexity of the first convolution layer on both graphs is 
\( O \left(m_t d_{\text{max}} + n_t {d_{\text{max}}}^2\right) \),
where $d_{\text{max}}$ is
the maximum size that a layer may have.
%In our model, \( |F_{\text{max}}| \) is the input feature size of the first layer, which is \( |F_{\text{input}}| \). Following this,
\item
The outputs of the first convolution layer for both graphs are aggregated, which incurs an additional complexity of
 \( O\left(n_t {d_{\text{max}}} \right) \) for element-wise operations or \( O \left(n_t {d_{\text{max}}}^2 \right) \) for concatenation and dimensionality reduction.
\item
The second convolution layer, which maps the aggregated vectors, contributes a complexity of
\( O \left(m_t {d_{\text{max}}} + n_t {d_{\text{max}}}^2 \right) \).
\end{itemize}
Thus, the total time complexity of AugSS-GNN is \( O \left(m_t {d_{\text{max}}} + n_t {d_{\text{max}}}^2 \right)  \), which is equal to that of a standard GNN model. The only computational overhead arises from the preprocessing step to construct the content graph, which involves computing a similarity measure between every pair of nodes, resulting in a complexity of
\( O \left(d  n_t^2 \right) \). However, these similarities are computed and stored once, eliminating the need for recalculation in subsequent model runs. Furthermore, in inductive settings, when a new node enters the graph, its similarity with other nodes only needs to be computed once, which has a complexity of \( O \left(d  n_t \right)  \). 
%Thiw demonstrates that the proposed model, while incorporating additional aggregation steps and dynamic trainable weights, maintains the same order of computational efficiency as traditional GNNs.

% Each convolution layer has \( O \left(|F_{\text{max}}|^2 \right)\) parameters. There are also 2 additional scalar parameters for aggregation. 
%The number of parameters of AugSS-GNN is primarily determined by two graph convolution layers and an aggregation step, 
%where each one has \( O \left( {d_{\text{max}}}^2 \right)\) parameters.
%Therefore, parameter complexity of Aug-SS  is \( O\left( {d_{\text{max}}}^2 \right) \), which is the same as standard GNNs.
The number of parameters of AugSS-GNN is primarily determined by two graph convolution layers and an aggregation step, each with
$ O \left( {d_{\text{max}}}^2 \right)$ parameters. Therefore, the parameter complexity of AugSS-GNN is
$ O\left( {d_{\text{max}}}^2 \right) $ , which is the same as standard GNNs.

\section{Experiments}
\label{sec:exp}

In this section, 
by conducting experiments over several real-world datasets, 
we show that our content augmentation techniques considerably improve the performance of GNNs.
We choose three well-known GNN models to improve by our augmentation techniques:
GCN \cite{DBLP:conf/iclr/KipfW17}, GAT \cite{DBLP:conf/iclr/VelickovicCCRLB18} and GATv2 \cite{brody2022how}.
Moreover, we compare the augmented GNNs against 
two well-known GNN augmentation methods, namely
LinkX \cite{lim2021large} and skip-connection \cite{you2020design}.
Furthermore, 
in order to evaluate the impact of nodes' features, we compare our methods against an MLP that only exploits nodes' content information. Additionally, we also consider using embeddings generated by DeepWalk~\cite{DeepWalk-14} as the input feature vectors for GNNs. The experiments are conducted on a Colab T4 GPU with 16 GB VRAM available for computation.

\subsection{Datasets}

To evaluate our models, we utilize five widely used datasets: Cora~\cite{yang2016revisiting}, CiteSeer~\cite{yang2016revisiting}, DBLP~\cite{bojchevski2017deep}, BlogCatalog~\cite{DBLP:journals/corr/abs-2009-00826} and Wiki~\cite{DBLP:journals/corr/abs-2009-00826}. The specifications of these real-world datasets are summarized in Table~\ref{tab:Datasets}.
We select these datasets for evaluation due to their inclusion of both graph-based and textual content.
%, a crucial characteristic for addressing real-world scenarios.
We partition each dataset into training, validation, and test sets.

\begin{table}[H]
\centering
\caption{Summary of our real-world datasets.}
\label{tab:Datasets}
\begin{tabular}{|c|c|c|c|c|c|c|} 
\hline
Dataset     & \#nodes & \#edges & Avg. degree & \multicolumn{1}{l|}{Avg. clustering coefficient} & \#attributes & \#classes  \\ 
\hline
Cora        & 2708    & 5278    & 3.89       & 0.246175                                     & 1433         & 7          \\
CiteSeer    & 3312    & 4732    & 2.85       & 0.144651                                     & 3703         & 6          \\
Wiki        & 2405    & 17981   & 14.95      & 0.323781                                     & 4973         & 17         \\
DBLP        & 17716   & 105734  & 11.92      & 0.134439                                     & 1639         & 4          \\
BlogCatalog & 5196    & 343486  & 132.2      & 0.122371                                     & 8189         & 6          \\
\hline
\end{tabular}
\end{table}

\subsection{Evaluation measures}

We adopt accuracy, macro-F1 and macro AUC-ROC as the evaluation criteria. 
%These metrics provide valuable insights into the performance of our model across different classification tasks.
Accuracy measures the proportion of correctly classified instances.
Macro-F1 accounts for both precision and recall, considering class imbalances and providing a balanced assessment of the model's performance.
%By utilizing these two metrics, we ensure a comprehensive evaluation that considers both overall accuracy and the model's ability to handle imbalanced class distributions.
%, enabling us to make well-informed decisions and draw meaningful conclusions about our model's effectiveness.
They are formally defined as follows.

\begin{equation}
	{accuracy} = \frac{\text{number of correctly classified test examples}}{\text{total number of test examples}} \times 100.
\end{equation}

\begin{equation}
	precision_{class (k)} = \frac{TP_{class (k)}}{TP_{class (k)} + FP_{class (k)}},
\end{equation}
where \(TP_{\text{class}(k)}\) represents the number of correctly predicted examples in class \(k\)
and \(FP_{\text{class}(k)}\) indicates the number of examples predicted as class \(k\)
but belong to other classes.

\begin{equation}
	recall_{class (k)} =\frac{TP_{class (k)}}{TP_{class (k)} + FN_{class (k)}},
\end{equation}
where \(FN_{\text{class}(k)}\) is defined as the number of examples in class \(k\) that are predicted as belonging to other classes.

%\begin{equation}
%    Accuracy = \frac{TP + TN}{TP+TN+FT+FN}   .
%\end{equation}
%
%\begin{equation}
%    Precision =\frac{TP}{TP + FP}.
%\end{equation}
%
%\begin{equation}
%    Recall =\frac{TP}{TP + FN}.
%\end{equation}
%
\begin{equation}
	F1_{class (k)} =\frac{2 \times precision_{class (k)} \times recall_{class (k)}}{precision_{class (k)} + recall_{class (k)}},
\end{equation}
and
\begin{equation}
	macro\text{-}F1 = \frac{1}{c} \times \sum_{k=1}^{c} F1_{class (k)},
\end{equation}
where $c$ is the number of classes.

%Our next evaluation criterion is  macro AUC-ROC. 
Macro AUC-ROC provides an aggregate measure of performance across all classes by calculating the Area Under the Curve (AUC) of the Receiver Operating Characteristic (ROC) curve for each class, then averaging these values. The AUC-ROC score represents the probability that a randomly chosen positive instance is ranked higher than a randomly chosen negative instance, with the ROC curve plotting the true positive rate (TPR) against the false positive rate (FPR) at various threshold settings. The AUC-ROC for a single class \(k\) is defined as follows: 
\begin{equation}
	{AUC-ROC}_{{class}(k)} = \int_{0}^{1} {TPR}_{{class}(k)}\left({FPR}_{{class}(k)}\right) \, d{FPR}_{{class}(k)},
\end{equation}
where \({TPR}_{{class}(k)}\) is the true positive rate for class \(k\) and \({FPR}_{{class}(k)}\) is the false positive rate for class \(k\). In this formula, \({TPR}\) is treated as a function of \({FPR}\), and \({FPR}\) serves as the variable of integration.
Macro AUC-ROC is then calculated as follows:
%\begin{equation}
% 	{AUC-ROC}_{{class}(k)} = \int_{0}^{1} {TPR}_{{class}(k)}({FPR}) \, d{FPR},
%  \end{equation} 
% where \({TPR}_{{class}(k)}\) is the true positive rate for class \(k\) and \({FPR}_{{class}(k)}\) is the false positive rate for class \(k\).
% Macro AUC-ROC is then calculated as follows: 
\begin{equation} 	
macro\text{ }AUC-ROC = \frac{1}{c} \sum_{k=1}^{c} {AUC-ROC}_{{class}(k)}, 
\end{equation}
 where \(c\) is the number of classes.

\subsection{Results}

In this section, we present the results of our experiments, 
in both supervised and semi-supervised settings.

\subsubsection{Supervised content augmentation}
\hfil\\
In the supervised setting, we split each dataset in half for training and testing. We report accuracy, macro-F1 and macro AUC-ROC scores for all the examined methods.
We run each model for $10$ times and report the average results as well as the standard deviations. 
%These reported scores are based on the results of $10$ runs of the models,
%where the average accuracy and macro-F1 scores as well as the standard deviations are provided in the tables.

The accuracy results presented in Table~\ref{tab:Expriment-Aug1--ACC} show that   our augmentation methods usually outperform the other methods.
 Notably, in specific datasets such as DBLP and BLOGCAT, the enhancements achieved by AugS are considerable. Furthermore, the macro-F1 scores detailed in Table~\ref{tab:Experiment-Aug1-F1} reveal noticeable improvements attributable to AugS, underscoring its effectiveness in handling imbalanced datasets with classes that have few instances.  In all the tables, rows marked with * indicate that the model runs with self-loops.
To further confirm the high performance of AugS, 
we report the macro AUC-ROC scores in Table~\ref{tab:Datasets_auc_roc}.

\begin{table}[H]
	\centering
	\caption{Comparing accuracy scores of the models in the supervised setting.\label{tab:Expriment-Aug1--ACC}}
	\label{tab:Expriment-Aug1--ACC}
	\scalebox{0.85}{
	\begin{tabular}{c|c|c|c|c|c}
		Model              & CORA                  & CITESEER              & WIKI                  & DBLP                  & BLOGCAT                \\ 
		\hline
		GCN                     & 86.92±0.26            & 73.32±0.08            & 74.62±0.15            & 72.85±0.72            & 70.94±4.25             \\
		GCN*                    & 85.39±0.03            & 76.09±0.08            & 75.45±0.23            & 80.21±0.02            & 71.88±2.97             \\
		AugS-GCN                & 87.41±0.55            & 72.44±0.38            & \textbf{ 76.14±0.94 } & 79.46±0.95            & 85.79±1.8              \\
		AugS-GCN*               & 86.70±0.70            & 75.78±0.26            & 75.51±1.12            & 83.09±0.13            & \textbf{ 86.04±1.33 }  \\ 
		\hline
		GAT                     & 85.11±0.00            & 75.83±0.13            & 66.27±1.19            & 73.25±0.09            & 59.05±0.41             \\
		GAT*                    & 85.27±0.11            & \textbf{ 77.12±0.11 } & 73.47±0.74            & 72.83±0.019           & 64.62±0.50             \\
		AugS-GAT                & \textbf{ 87.65±0.46 } & 74.76±0.52            & 73.01±1.72            & \textbf{ 84.58±0.12 } & 67.33±0.68             \\
		AugS- GAT *             & 86.17±0.04            & 76.34±0.43            & 73.67±0.60            & 83.00±0.19            & 72.52±0.96             \\ 
		\hline
		GATv2                   & 87.10±0.11            & 75.90±0.14            & 72.96±0.58            & 67.24±0.81            & 61.78±0.80             \\
		GATv2*                  & 86.07±0.03            & 76.36±0.09            & 74.44±0.08            & 79.90±0.62            & 62.29±0.33             \\
		AugS-GATv2              & 87.52±0.37            & 75.32±0.16            & 73.40±1.10            & 82.18±0.24            & 68.56±0.62             \\
		AugS- GATv2 *           & 86.01±0.03            & 74.8±0.22             & 75.78±0.50            & 83.69±0.25            & 70.94±0.81             \\ 
		\hline
		DeepWalk-GCN            & 80.56±0.01            & 62.40±0.12            & 62.40±0.12            & 80.73±0.02            & 63.10±0.12             \\
		DeepWalk-GAT            & 81.27±0.05            & 62.63±0.06            & 62.63±0.06            & 80.49±0.01            & 61.88±0.05             \\
		DeepWalk-GATv2          & 80.27±0.11            & 62.82±0.12            & 62.85±0.12            & 80.58±0.04            & 64.57±0.05             \\ 
		\hline
		MLP                     & 71.8±2.54             & 69.87±1.52            & 75.32±1.88            & 75.76±1.15            & 85.47±0.50             \\ 
		\hline
		LinkX                   & 78.87±0.41            & 58.3±0.45             & 68.05±0.42            & 80.04±0.40            & 72.63±1.21             \\ 
		\hline
		Skip-connection-GCN   & 86.51±0.14            & 72.37±0.10            & 74.44±0.18            & 81.54±0.01            & 85.64±0.03             \\
		Skip-connection-GAT   & 85.90±0.25            & 72.91±0.28            & 74.89±0.57            & 82.48±0.01            & 77.63±0.50             \\
		Skip-connection-GATv2 & 86.93±0.16            & 72.41±0.36            & 73.47±0.31            & 82.17±0.02            & 75.26±0.65            
	\end{tabular}
}
\end{table}

\begin{table}[H]
\centering
\caption{Comparing macro F1-scores of the models in the supervised setting.}
\label{tab:Experiment-Aug1-F1}
\scalebox{0.85}{
\begin{tabular}{c|c|c|c|c|c}
Model             & CORA                  & CITESEER              & WIKI                  & DBLP                  & BLOGCAT                \\ 
\hline
GCN                     & 84.87±0.12            & 69.33±0.07            & 53.72±1.28            & 61.12±0.39            & 68.31±7.36             \\
GCN*                    & 83.50±0.03            & 71.32±0.14            & 64.00±0.53            & 64.60±0.06            & 69.62±4.68             \\
AugS-GCN                & \textbf{ 86.48±0.50 } & 70.10±0.11            & 63.89±3.01            & 60.10±0.23            & 76.97±0.60             \\
AugS-GCN*               & 84.19±0.07            & 69.02±0.63            & 59.40±3.01            & 78.73±0.22            & \textbf{ 85.63±1.34 }  \\ 
\hline
GAT                     & 82.47±0.03            & 70.81±0.10            & 55.03±2.20            & 50.20±0.054           & 60.70±4.91             \\
GAT*                    & 83.75±0.14            & 71.59±0.18            & 57.17±2.29            & 54.07±0.13            & 58.19±0.40             \\
AugS-GAT                & 86.22±0.65            & 70.11±0.37            & 58.79±2.14            & 77.78±0.41            & 67.22±0.63             \\
AugS-GAT*               & 84.52±0.51            & 71.48±0.79            & 58.90±3.15            & \textbf{ 81.34±0.15 } & 71.88±0.98             \\ 
\hline
GATv2                   & 84.87±0.12            & 69.49±0.10            & 51.47±2.30            & 65.47±0.39            & 61.22±0.87             \\
GATv2*                  & 84.08±0.09            & 71.50±0.17            & 57.88±3.12            & 73.63±0.40            & 61.615±0.37            \\
AugS-GATv2              & 85.80±0.62            & \textbf{ 72.48±0.56 } & 54.58±1.18            & 77.78±0.41            & 68.40±0.65             \\
AugS-GATv2*             & 84.10±0.31            & 70.22±0.88            & 58.90±4.16            & 80.12±0.36            & 70.57±0.78             \\ 
\hline
DeepWalk-GCN            & 79.18±0.00            & 56.38±0.04            & 42.62±0.06            & 75.82±0.01            & 63.40±0.05             \\
DeepWalk-GAT            & 79.29±0.12            & 52.07±0.13            & 42.34±0.03            & 75.90±0.00            & 60.35±0.06             \\
DeepWalk-GATv2          & 79.37±0.08            & 53.65±0.15            & 43.20±0.07            & 75.88±0.07            & 61.00±0.12             \\ 
\hline
MLP                     & 68.74±3.75            & 66.89±1.88            & 66.17±3.80            & 70.34±2.30            & 85.25±1.1              \\ 
\hline
LinkX                   & 79.23±0.20            & 56.80±0.50            & \textbf{ 68.19±0.39 } & 78.68±1.1             & 74.2±0.9               \\ 
\hline
Skip-connection-GCN   & 85.59±0.18            & 69.01±0.01            & 63.19±0.13            & 76.37±0.09            & 85.04±0.43             \\
Skip-connection-GAT   & 84.39±0.25            & 68.85±0.34            & 64.83±0.68            & 76.75±0.02            & 77.57±0.49             \\
Skip-connection-GATv2 & 85.42±0.17            & 69.36±0.18            & 66.16±0.21            & 75.68±0.04            & 77.09±0.62            
\end{tabular}
}
\end{table}

\begin{table}[H]
\centering
\caption{Comparing macro AUC-ROC scores of the models in the supervised setting.}
\label{tab:Datasets_auc_roc}
\scalebox{0.85}{
\begin{tabular}{c|c|c|c|c|c}
Model        & CORA                  & CITESEER              & WIKI                  & DBLP                  & BLOGCAT                \\ 
\hline
GCN                   & 97.15±0.14                & 88.17±0.07            & 92.63±0.01            & 91.92±0.00            & 90.95±0.24             \\
GCN*                    & 96.95±0.19            & 92.50±0.17            & 92.05±0.01            & 95.11±0.00            & 91.65±0.62             \\
AugS-GCN                & 98.22±0.00            & 90.12±0.61            & 94.77±0.00            & 92.91±0.45            & 95.96±0.17             \\
AugS-GCN*               & 97.05±0.01            & \textbf{ 92.57±0.11 } & 93.40±0.01            & 95.55±0.56            & \textbf{ 98.76±0.21 }  \\ 
\hline
GAT                     & 97.42±0.04            & 91.66±0.17            & 87.54±0.97            & 91.79±0.00            & 86.45±0.16             \\
GAT*                    & 96.24±0.10            & 92.48±0.05            & 88.73±0.08            & 92.08±0.03            & 85.47±0.29             \\
AugS-GAT                & 98.21±0.04            & 89.56±0.14            & 88.50±0.07            & \textbf{ 95.77±0.08 } & 89.56±0.13             \\
AugS-GAT *              & 96.86±0.05            & 91.95±0.02            & 88.98±0.08            & 95.03±0.12            & 91.29±0.36             \\ 
\hline
GATv2                   & 97.94±0.01            & 92.02±0.04            & 92.41±0.15            & 89.90±0.33            & 88.31±0.17             \\
GATv2*                  & 97.09±0.00            & 92.41±0.01            & 91.02±0.12            & 92.38±1.20            & 86.60±0.42             \\
AugS-GATv2              & \textbf{ 98.25±0.02 } & 91.98±0.01            & 93.05±0.07            & 95.58±0.00            & 90.92±0.44             \\
AugS-GATv2 *            & 96.58±0.05            & 92.11±0.03            & \textbf{ 94.81±0.21 } & 95.74±0.29            & 90.86±0.32             \\ 
\hline
DeepWalk-GCN            & 96.35±0.00            & 84.57±0.01            & 88.03±0.00            & 93.24±0.00            & 90.53±0.01             \\
DeepWalk-GAT            & 96.07±0.00            & 82.56±0.00            & 88.84±0.00            & 92.88±0.00            & 88.40±0.00             \\
DeepWalk-GATv2          & 96.07±0.00            & 82.49±0.00            & 89.33±0.05            & 93.18±0.00            & 89.46±0.00             \\ 
\hline
MLP                     & 93.02±0.02            & 89.75±0.05            & 92.17±1.77            & 91.67±0.04            & 98.52±0.10             \\ 
\hline
LinkX                   & 92.23±0.09            & 84.7±0.07             & 85.22±0.15            & 91.02±0.32            & 92.51±0.05             \\ 
\hline
Skip-connection-GCN   & 97.77±0.00            & 90.61±0.01            & 94.52±0.06            & 93.67±0.00            & 97.3±0.010             \\
Skip-connection-GAT   & 96.75±0.04            & 90.37±0.05            & 94.22±0.15            & 93.23±0.01            & 93.72±0.00             \\
Skip-connection-GATv2 & 97.63±0.01              & 89.87±0.10            & 94.48±0.13            & 93.56±0.00            & 92.62±0.04            
\end{tabular}
}
\end{table}

We see in our experiments that
having a sufficient number of nodes and features enables supervised methods to relay heavily on node features, particularly when the average degree exceeds 10. Additionally, incorporating self-loops prevents the neural network from losing the initial information of the nodes. Our experiments generally demonstrate that self-loops enhance the performance of the model, except in the Cora dataset, which does not exhibit the same behavior as other datasets. In Cora, we encounter a sparse network with a low number of features, leading to improved performance from the graph structure itself. Therefore, in this case, focusing extra attention on initial features does not enhance the model, resulting in no performance improvement with self-loops in Cora.

%We observed improved performance with self-connections in most cases.
AugS, especially in the auto-decoder part, requires substantial training data. Consequently, although it enhances performance in most cases, it exhibits weaknesses over Citeseer, which has a low number of nodes and features, as well as the lowest edge degree among our datasets.
On the other hand, over datasets with rich node features and a sufficient number of nodes, such as BlogCatalog, AugS shows significant performance improvements.
%Over Citeseer, despite increasing the training data to 50 percent of the nodes to form a supervised setting, AugS's performance does not show an improvement.
 This weaker performance of AugS over Citeseer
 is  related to the insufficient number of nodes and training data, which negatively impact its functionality.
% Therefore, in Citeseer, even with 50 percent of the training size, the model could still be considered as operating in a semi-supervised or supervised mode, capable of training with fewer data points while achieving better generalization.

%Overall, in most of our experiments, our initial concept, when combined with other networks, has improved their functionality. Compared to other types of GNN (Graph Neural Network) models, ours has shown superior performance.

\begin{figure}[H]
    \centering
    \includegraphics[width=\linewidth]{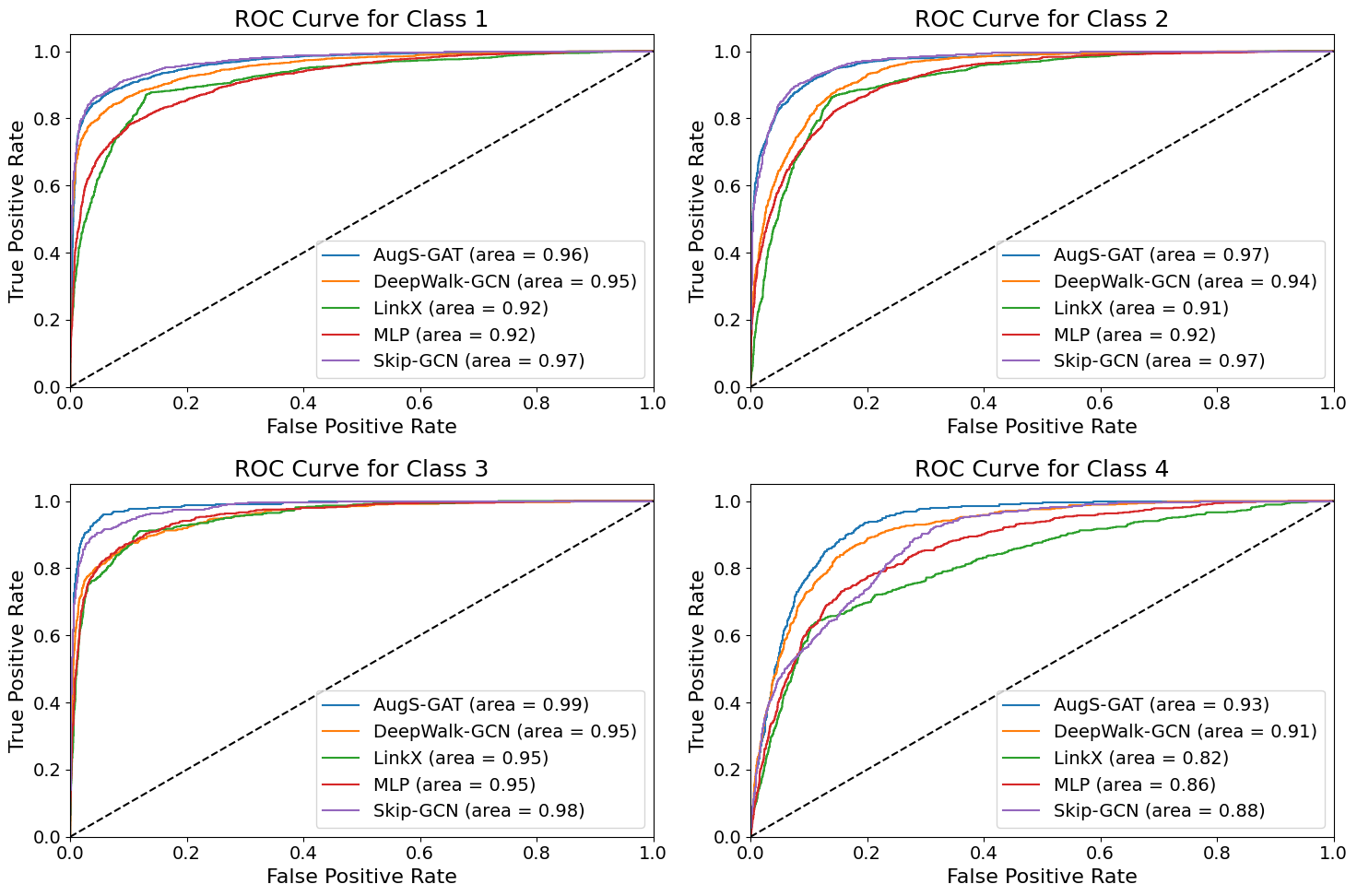}
    \caption{ROC curves of AugS-GAT and the best configurations of the baseline models, in the supervised setting.}
    \label{fig:ROC_Augs_figure}
\end{figure}

In order to examine the performance of AugS over different classes,
we choose the DBLP dataset and in
Figure~\ref{fig:ROC_Augs_figure}, present the ROC scores of AugS-GAT,
%  on the DBLP dataset, illustrating the model's behavior
   across different classes within this dataset. 
 %To compare model performance, we selected our best model, GAT-Augs, alongside 
We compare AugS-GAT against four other models:
the best skip-connection model (which is skip-connection-GCN),
the best DeepWalk model (which is DeepWalk-GCN), MLP, and LinkX.
As can be seen, in classes 1 and 2, AugS-GAT and skip-connection-GCN are the best methods. However, in classes 3 and 4, AugS-GAT shows significantly better results than the other methods. It is important to note that although the skip-connection model performs closely to our model over three classes, 
it does not show a good performance in class 4. This indicates that the skip-connection model may 
not perform well over minority classes.
%not consistently detect different classes effectively. 
Conversely, AugS-GAT exhibits high performances across all classes, demonstrating its robustness even in classes 3 and 4, which have a minority population in our dataset.

\begin{table}[h]
	\caption{Class distribution of the DBLP dataset.}
	\centering
	\label{tab:DBLP_CAT}
	\begin{tabular}{|c|c|} 
		\hline
		\textbf{class} & \textbf{\#examples}  \\ 
		\hline
		1              & 7920               \\
		2              & 5645               \\
		3              & 1928               \\
		4              & 2169               \\
		\hline
	\end{tabular}
\end{table}

\subsubsection{Semi-supervised content augmentation}
\hfil\\
In this section, we assess the performance of AugSS-GNN, within a semi-supervised framework. To provide a limited amount of training data for the models,
we select $20$ labeled samples per class from our datasets
and randomly sample $500$ data points for validation.
Additionally, we allocate $1000$ data points for evaluating the models.
In the case of the wiki dataset, some classes have fewer than $20$ samples,
prompting us to select just $5$ training samples from these classes.
We present accuracy (Table~\ref{tab:Expriment-Aug2-F1}), 
macro-F1 scores (Table~\ref{tab:Experiment-Aug2-ACC})
and AUC-ROC scores
 (Table~\ref{tab:Datasets_auc_roc_AugSS}),
for the examined models.
These reported scores are based on the results of $10$ runs of the models,
where the average accuracy, average macro-F1 and average AUC-ROC  as well as the standard deviations are provided in the tables.

Our experiments illustrate the impact of incorporating a content graph into the training process of graph neural networks.
By using AugSS,
across various datasets,
we observe noticeable improvements in the accuracy and F1-scores of the base GNN models, ranging from approximately $2\%$ to $14\%$.
Therefore, AugSS considerably enhances the performance of GNNs,
when a limited amount of labeled data is available during the training phase.
This improvement in the performance of GNN models is due to 
the role of nodes' content information in their discrimination power,
which is better modeled and reflected to the final classifier,
using our content augmentation method.
%This demonstrates the effectiveness of our proposed method in enhancing the training of graph neural networks.

\definecolor{Shark}{rgb}{0.121,0.121,0.125}
\definecolor{MineShaft}{rgb}{0.129,0.129,0.129}
\begin{table}[H]
\centering
\caption{Comparing  accuracy scores of the models, in the semi-supervised setting.}
\label{tab:Expriment-Aug2-F1}
\scalebox{0.85}{
\begin{tblr}{
  cells = {c},
  cell{13}{2} = {fg=Shark},
  cell{13}{4} = {fg=MineShaft},
  cell{14}{4} = {fg=MineShaft},
  cell{15}{4} = {fg=MineShaft},
  vline{2-6} = {-}{},
  hline{2,4,6,8,11-13} = {-}{},
}
Model               & CORA                 & CITESEER             & WIKI                  & DBLP                  & BLOGCAT               \\
GCN                      & 80.81 ± 0.60         & 69.32 ± 0.69         & 68.65 ± 1.26          & 74.92 ± 1.03          & 75.38 ± 0.83          \\
AugSS-GCN                 & \textbf{82.48 ± 0.9} & \textbf{72.85 ±0.55} & \textbf{72.61 ± 0.84} & \textbf{77.31 ± 0.65} & 72.87 ± 0.9           \\
GAT                      & 81.17 ± 1.21         & 68.14 ± 1.51         & 52.89 ± 3.92          & 75.4 ± 0.96           & 63.54 ± 2.5           \\
AugSS-GAT                 & 82.14 ± 0.71         & 72.46 ± 0.53         & 59.5 ± 0.74           & 76.33 ± 0.71          & 67.6 ± 2.83           \\
GATv2                    & 80.64 ± 0.62         & 68.2 ± 1.11          & 50.61 ± 3.31          & 74.85 ± 1.36          & 67.57 ± 2.1           \\
AugSS-GATv2               & 81.62 ± 0.42         & 72.50 ± 0.69         & 59.7 ± 1.55           & 76.42 ± 0.82          & \textbf{81.28 ± 1.33} \\
DeepWalk-GCN             & 74.50 ± 0.00         & 45.56 ± 0.19         & 54.81 ± 0.32          & 70.44 ± 0.11          & 60.76 ± 0.05          \\
DeepWalk-GAT             & 71.26 ± 0.05         & 42.36 ± 0.18         & 49.86 ± 0.37          & 72.74 ± 0.05          & 56.84 ± 0.05          \\
DeepWalk-GATv2           & 80.44 ± 0.16         & 44.44 ± 0.05         & 52.90 ± 0.12          & 70.94 ± 0.17          & 57.00 ± 0.13          \\
MLP                      & 54.32±0. 98          & 58.01±0. 83          & 69.70±0. 32           & 47.93± 1.09           & 68.71 ± 3.96          \\
LinkX                    & 50.45 ± 0.76         & 36.45 ± 0.37         & 37.30 ± 0.72          & 33.31 ± 7.40          & 57.86 ± 0.47          \\
Skip-connection-GCN   & 
  79.30 ± 0.19
     & 65.06 ± 0.13         & 
  63.86 ± 0.45
      & 69.10 ± 0.07          & 78.62 ± 0.23          \\
Skip-connection-GAT   & 71.24 ± 0.47         & 65.04 ± 0.61         & 
  61.56 ± 0.32
      & 73.12 ± 0.71          & 40.16 ± 1.06          \\
Skip-connection-GATv2 & 73.72 ± 0.85         & 62.60 ± 0.98         & 
  60.54 ± 0.32
      & 72.38 ± 1.36          & 36.34 ± 0.47          
\end{tblr}
}
\end{table}

\definecolor{Shark}{rgb}{0.121,0.121,0.125}
\begin{table}[H]
\centering
\caption{Comparing F1-scores of the models, in the semi-supervised setting.
\label{tab:Experiment-Aug2-ACC}}
\scalebox{0.85}{
\begin{tblr}{
  cells = {c},
  cell{2}{2} = {fg=Shark},
  cell{2}{3} = {fg=Shark},
  cell{2}{4} = {fg=Shark},
  cell{2}{5} = {fg=Shark},
  cell{2}{6} = {fg=Shark},
  cell{3}{2} = {fg=Shark},
  cell{3}{3} = {fg=Shark},
  cell{3}{4} = {fg=Shark},
  cell{3}{5} = {fg=Shark},
  cell{3}{6} = {fg=Shark},
  cell{4}{2} = {fg=Shark},
  cell{4}{3} = {fg=Shark},
  cell{4}{4} = {fg=Shark},
  cell{4}{5} = {fg=Shark},
  cell{4}{6} = {fg=Shark},
  cell{5}{2} = {fg=Shark},
  cell{5}{3} = {fg=Shark},
  cell{5}{4} = {fg=Shark},
  cell{5}{5} = {fg=Shark},
  cell{5}{6} = {fg=Shark},
  cell{6}{2} = {fg=Shark},
  cell{6}{3} = {fg=Shark},
  cell{6}{4} = {fg=Shark},
  cell{6}{5} = {fg=Shark},
  cell{6}{6} = {fg=Shark},
  cell{7}{2} = {fg=Shark},
  cell{7}{3} = {fg=Shark},
  cell{7}{4} = {fg=Shark},
  cell{7}{5} = {fg=Shark},
  cell{7}{6} = {fg=Shark},
  cell{8}{2} = {fg=Shark},
  cell{8}{3} = {fg=Shark},
  cell{8}{4} = {fg=Shark},
  cell{8}{5} = {fg=Shark},
  cell{8}{6} = {fg=Shark},
  cell{9}{2} = {fg=Shark},
  cell{9}{3} = {fg=Shark},
  cell{9}{4} = {fg=Shark},
  cell{9}{5} = {fg=Shark},
  cell{9}{6} = {fg=Shark},
  cell{10}{2} = {fg=Shark},
  cell{10}{3} = {fg=Shark},
  cell{10}{4} = {fg=Shark},
  cell{10}{5} = {fg=Shark},
  cell{10}{6} = {fg=Shark},
  cell{11}{2} = {fg=Shark},
  cell{11}{3} = {fg=Shark},
  cell{11}{4} = {fg=Shark},
  cell{11}{5} = {fg=Shark},
  cell{11}{6} = {fg=Shark},
  cell{12}{2} = {fg=Shark},
  cell{12}{3} = {fg=Shark},
  cell{12}{4} = {fg=Shark},
  cell{12}{5} = {fg=Shark},
  cell{12}{6} = {fg=Shark},
  cell{13}{2} = {fg=Shark},
  cell{13}{3} = {fg=Shark},
  cell{13}{4} = {fg=Shark},
  cell{13}{5} = {fg=Shark},
  cell{13}{6} = {fg=Shark},
  cell{14}{2} = {fg=Shark},
  cell{14}{3} = {fg=Shark},
  cell{14}{4} = {fg=Shark},
  cell{14}{5} = {fg=Shark},
  cell{14}{6} = {fg=Shark},
  cell{15}{2} = {fg=Shark},
  cell{15}{3} = {fg=Shark},
  cell{15}{4} = {fg=Shark},
  cell{15}{5} = {fg=Shark},
  cell{15}{6} = {fg=Shark},
  vline{2-6} = {-}{},
  hline{2,4,6,8,11-13} = {-}{},
}
					   Model
                       & CORA                                                              & CITESEER                                                          & WIKI                 & DBLP                                                              & BLOGCAT                                                           \\
GCN                    & 79.91 ± 0.6                                                      & 65.62~± 0.9                                                       & 58.77~± 1.6          & 70.94~± 0.9                                                       & 74.70~± 0.9                                                       \\
AugSS-GCN              & \textbf{\textbf{81.35~}}\textbf{\textbf{±~}}\textbf{\textbf{0.9}} & 69.11~±~0.6                                                       & 62.78~±~1.1          & \textbf{\textbf{74.03~}}\textbf{\textbf{±~}}\textbf{\textbf{0.8}} & 71.99~±~1.0                                                       \\
GAT                    & 80.36~±~1.0                                                       & 68.45~±~0.0                                                       & 43.29~±~2.9          & 72.10~±~0.8                                                       & 62.99~±~2.2                                                       \\
AugSS-GAT              & 81.03~±~0.9                                                       & \textbf{\textbf{70.45~}}\textbf{\textbf{±~}}\textbf{\textbf{2.1}} & 48.85~±~1.5          & 72.81~±~0.8                                                       & 66.79~±~2.6                                                       \\
GATv2                  & 79.74~±~0.6                                                       & 64.62~±~0.9                                                       & 41.60~±~2.1          & 71.87~±~1.3                                                       & 66.41~±~2.3                                                       \\
AugSS-GATv2            & 80.65~±~0.6                                                       & 67.78~±~0.5                                                       & 47.62~±~1.7          & 72.94~±~1.2                                                       & \textbf{\textbf{80.74~}}\textbf{\textbf{±~}}\textbf{\textbf{1.4}} \\
Deep walk-GCN          & 73.61 ± 0.02                                                      & 42.80 ± 0.18                                                      & 52.44 ± 0.28         & 65.57 ± 0.16                                                      & 58.80 ± 0.06                                                      \\
Deep walk-GAT          & 70.32 ± 0.02                                                      & 40.32 ± 0.17                                                      & 45.45 ± 0.42         & 67.81 ± 0.07                                                      & 55.46 ± 0.05                                                      \\
Deep walk-GATv2        & 78.91 ± 0.19                                                      & 42.48 ± 0.10                                                      & 49.05 ± 0.08         & 66.80 ± 0.18                                                      & 55.60 ± 0.11                                                      \\
MLP                    & 51.90 ± 1.4                                                       & 56.52± 0.96                                                       & \textbf{69.21± 0.45} & 45.08 ± 1.9                                                       & 67.77 ± 4.6                                                       \\
LinkX                  & 48.27 ± 1.00                                                      & 35.60 ± 1.99                                                      & 35.39 ± 0.24         & 31.75 ± 7.74                                                      & 57.88 ± 0.45                                                      \\
Skip-Connection-GCN   & 78.96 ± 0.18                                                      & 62.06 ± 0.12                                                      & 62.18 ± 0.43         & 66.86 ± 0.07                                                      & 77.80 ± 0.25                                                      \\
Skip-Connection-GAT   & 71.99 ± 0.49                                                      & 61.26 ± 0.56                                                      & 59.60 ± 0.44         & 70.59 ± 0.67                                                      & 39.06 ± 0.92                                                      \\
Skip-Connection-GATv2 & 73.35 ± 0.65                                                      & 59.94 ± 0.94                                                      & 58.99 ± 0.33         & 69.27 ± 1.49                                                      & 35.95 ± 0.55                                                      
\end{tblr}
}
\end{table}

We observe that the improvements in the DBLP and Wiki datasets are more significant compared to the Cora and Citeseer datasets. Additionally, the BlogCatalog dataset shows the greatest improvements among all five datasets. This could be attributed to the average degree of the datasets. The BlogCatalog dataset has a significantly higher average degree, indicating that it is less sparse than the other datasets. This increased connectivity makes it more likely for the effect of initial node features to diminish after several GNN layers. Therefore, incorporating more content information can lead to greater improvements in this case. A similar trend is observed when comparing the Wiki and DBLP datasets with Cora and Citeseer.

To demonstrate the impact of incorporating both content and structural information, we conducted experiments using an MLP that considers only content features as input, and standard GNNs that utilize DeepWalk features, which are structural features, as input. The results,
presented in Tables~\ref{tab:Expriment-Aug2-F1}, 
\ref{tab:Experiment-Aug2-ACC} and \ref{tab:Datasets_auc_roc_AugSS},
show that our proposed model, which combines content features with structural features, performs considerably better.

\definecolor{Shark}{rgb}{0.121,0.121,0.125}
\begin{table}
\centering
\caption{Comparing macro AUC-ROC scores of the models, in the semi-supervised setting.}
\label{tab:Datasets_auc_roc_AugSS}
\scalebox{0.85}{
\begin{tblr}{
  cells = {c},
  cell{2}{2} = {fg=Shark},
  cell{2}{3} = {fg=Shark},
  cell{2}{4} = {fg=Shark},
  cell{2}{5} = {fg=Shark},
  cell{2}{6} = {fg=Shark},
  cell{3}{2} = {fg=Shark},
  cell{3}{3} = {fg=Shark},
  cell{3}{4} = {fg=Shark},
  cell{3}{5} = {fg=Shark},
  cell{3}{6} = {fg=Shark},
  cell{4}{2} = {fg=Shark},
  cell{4}{3} = {fg=Shark},
  cell{4}{4} = {fg=Shark},
  cell{4}{5} = {fg=Shark},
  cell{4}{6} = {fg=Shark},
  cell{5}{3} = {fg=Shark},
  cell{5}{4} = {fg=Shark},
  cell{5}{5} = {fg=Shark},
  cell{5}{6} = {fg=Shark},
  cell{6}{2} = {fg=Shark},
  cell{6}{3} = {fg=Shark},
  cell{6}{4} = {fg=Shark},
  cell{6}{5} = {fg=Shark},
  cell{6}{6} = {fg=Shark},
  cell{7}{2} = {fg=Shark},
  cell{7}{3} = {fg=Shark},
  cell{7}{4} = {fg=Shark},
  cell{7}{5} = {fg=Shark},
  cell{7}{6} = {fg=Shark},
  cell{8}{2} = {fg=Shark},
  cell{8}{3} = {fg=Shark},
  cell{8}{4} = {fg=Shark},
  cell{8}{5} = {fg=Shark},
  cell{8}{6} = {fg=Shark},
  cell{9}{2} = {fg=Shark},
  cell{9}{3} = {fg=Shark},
  cell{9}{4} = {fg=Shark},
  cell{9}{5} = {fg=Shark},
  cell{9}{6} = {fg=Shark},
  cell{10}{2} = {fg=Shark},
  cell{10}{3} = {fg=Shark},
  cell{10}{4} = {fg=Shark},
  cell{10}{5} = {fg=Shark},
  cell{10}{6} = {fg=Shark},
  cell{11}{2} = {fg=Shark},
  cell{11}{3} = {fg=Shark},
  cell{11}{4} = {fg=Shark},
  cell{11}{5} = {fg=Shark},
  cell{11}{6} = {fg=Shark},
  cell{12}{2} = {fg=Shark},
  cell{12}{3} = {fg=Shark},
  cell{12}{4} = {fg=Shark},
  cell{12}{5} = {fg=Shark},
  cell{12}{6} = {fg=Shark},
  cell{13}{2} = {fg=Shark},
  cell{13}{3} = {fg=Shark},
  cell{13}{4} = {fg=Shark},
  cell{13}{5} = {fg=Shark},
  cell{13}{6} = {fg=Shark},
  cell{14}{2} = {fg=Shark},
  cell{14}{3} = {fg=Shark},
  cell{14}{4} = {fg=Shark},
  cell{14}{5} = {fg=Shark},
  cell{14}{6} = {fg=Shark},
  cell{15}{2} = {fg=Shark},
  cell{15}{3} = {fg=Shark},
  cell{15}{4} = {fg=Shark},
  cell{15}{5} = {fg=Shark},
  cell{15}{6} = {fg=Shark},
  vline{2-6} = {-}{},
  hline{2,4,6,8,11-13} = {-}{},
}
Model              & CORA                  & CITESEER             & WIKI                & DBLP                  & BLOGCAT               \\
GCN                       & 
  97.10 ± 0.33
      & 
  88.4 ± 0.01
      & 
  93.42 ± 0.0
     & 
  90.77 ± 0.06
      & 
  94.97 ± 0.14
      \\
AugS-GCN                  & 
  97.24 ± 0.26
      & \textbf{90.66 ± 0.0} & \textbf{94.8 ± 0.0} & \textbf{92.04 ± 0.38} & 
  94.97 ± 0.14~ ~~   \\
GAT                       & 
  97.12 ± 0.21
      & 
  88.30 ± 0.48
     & 
  81.97 ± 0.04
    & 
  90.60 ± 0.05
      & 
  94.43 ± 0.09
      \\
AugS-GAT                  & \textbf{97.26 ± 0.16} & 
  89.34 ± 0.85
     & 
  86.55 ± 0.03
    & 
  91.65 ± 0.01~
     & 
  90.59 ± 0.86
      \\
GATv2                     & 
  96.85 ± 0.27
      & 
  88.51 ± 0.01
     & 
  87.36 ± 0.07
    & 
  91.29 ± 0.04~
     & 
  91.02 ± 0.83
      \\
AugS-GATv2                & 
  96.86 ± 0.38
      & 
  90.28 ± 0.82
     & 
  89.61 ± 0.11
    & 
  92.01 ± 0.01
      & \textbf{95.62 ± 0.90} \\
DeepWalk-GCN              & 
  93.75 ± 0.00
      & 
  71.60 ± 0.03
     & 
  89.65 ± 0.01
    & 
  87.15 ± 0.02
      & 
  87.33 ± 0.01
      \\
DeepWalk-GAT              & 
  93.65 ± 0.02
      & 
  71.25 ± 0.03
     & 
  87.67 ± 0.01
    & 
  88.41 ± 0.01
      & 
  84.58 ± 0.02
      \\
DeepWalk-GATv2            & 
  95.05 ± 0.01
      & 
  71.03 ± 0.04
     & 
  88.88 ± 0.05
    & 
  87.85 ± 0.02
      & 
  83.10 ± 0.04
      \\
MLP                       & 
  85.94±0.02
        & 
  85.44±0.06
       & 
  92.60±0.03
      & 
  73.47±0.03
        & 
  92.75±0.11
        \\
LinkX                     & 
  83.21 ± 0.04~
     & 
  65.28 ± 1.12
     & 
  89.46 ± 0.03~
   & 
  62.26 ± 1.40~
     & 
  85.96 ± 0.15~
     \\
Skip-connection-GCN  & 
  96.73 ± 0.02~ ~~   & 
  87.57 ± 0.07
     & 
  93.82 ± 0.04
    & 
  90.38 ± 0.02
      & 
  94.40 ± 0.04
      \\
Skip-connection-GAT  & 94.45 ± 0.07          & 
  86.77 ± 0.21
     & 
  93.96 ± 0.05
    & 
  90.03 ± 0.07
      & 
  71.62 ± 0.55
      \\
Skip-connection-GATv2 & 
  94.28 ± 0.17
      & 
  86.64 ± 0.06
     & 
  93.42 ± 0.11
    & 
  90.77 ± 0.32
      & 
  69.83 ± 0.47
      
\end{tblr}
}
\end{table}

\begin{figure}
    \centering
    \includegraphics[width=\linewidth]{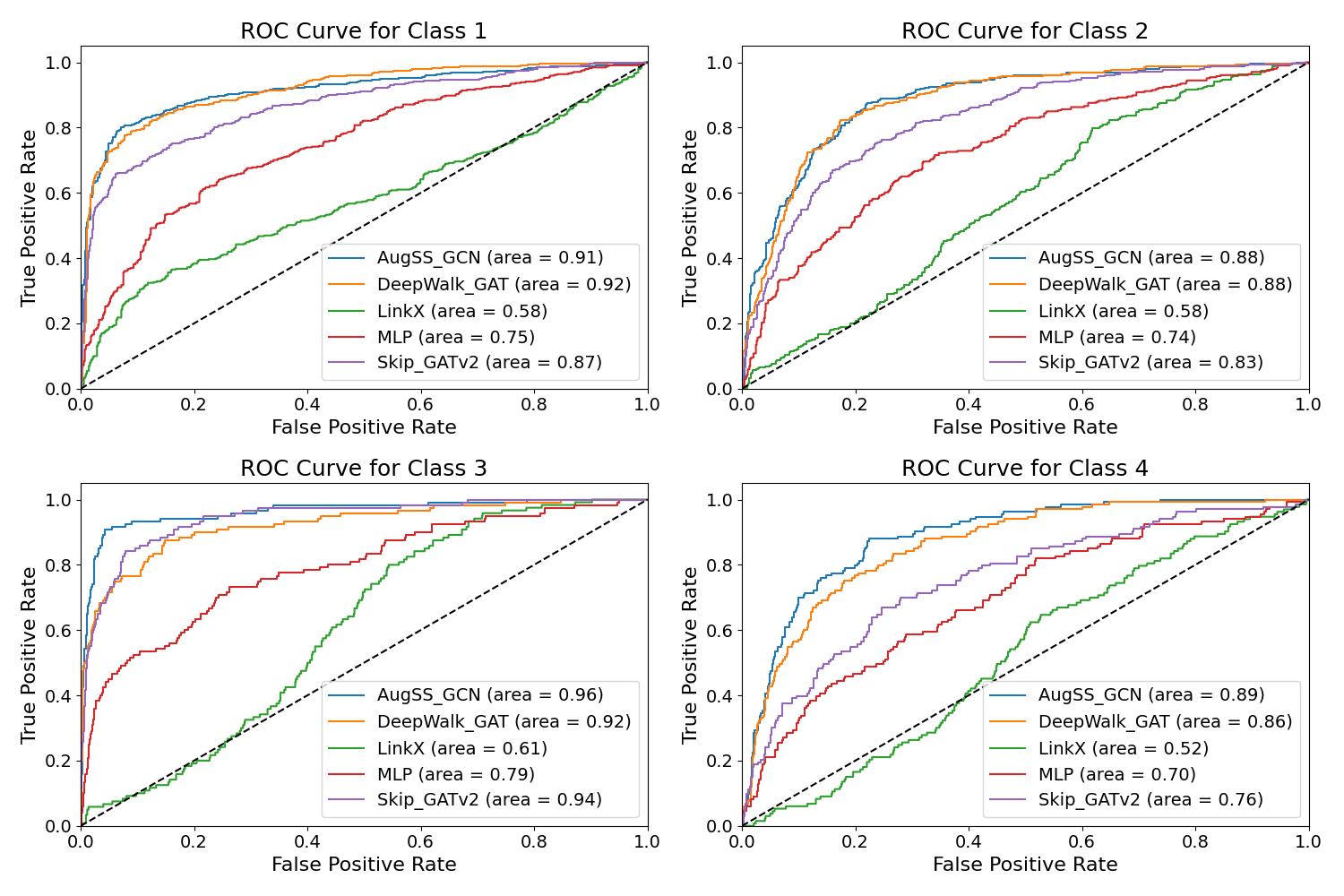}
    \caption{ROC curves of AugSS-GCN against the best configurations of the baseline models, in the semi-supervised setting.}
    \label{fig:ROC_Augss_figure2}
\end{figure}

Figure \ref{fig:ROC_Augss_figure2} presents the ROC curves of
the best configuration of AugSS (which is AugSS-GCN), against the best configurations of the baseline models DeepWalk, skip-connection, MLP, and LinkX,
for different classes of the DBLP dataset.
% The models evaluated include the best-performing AugSS (AugSS-GCN), DeepWalk, skip-connection, MLP, and LinkX models. As depicted,
As can be seen in the figure, AugSS-GCN consistently outperforms the other models across nearly all classes. The only exception is class 1, where DeepWalk-GAT achieves superior performance. While skip-connection-GATv2 exhibits strong performance in class 3, it does not perform well in  other classes. Both LinkX and MLP fail to yield satisfactory results in any of the classes. Moreover, DeepWalk-GAT demonstrates relatively good performance across all classes, particularly in classes 1 and 2, where its results are comparable to those of AugSS-GCN. The poor performance of MLP compared to the good performance of DeepWalk-GAT suggests that structural features are more critical than content features over the DBLP dataset. 
DeepWalk utilizes only structural features, MLP relies solely on content features, and AugSS-GCN, which combines both structural and content features, outperforms both models.

\FloatBarrier

\section{Conclusion}
\label{sec:conclusion}
%In this paper, we proposed two novel approaches that augment node embeddings with content-based embeddings at higher GNN layers. In AugS, for each node, a structural embedding and a content embedding are computed and combined using a combination layer to form the embedding of the node at a given layer. We presented techniques such as an unsupervised auto-encoder technique, which is used in supervised settings, and a content graph strategy, which is useful in semi-supervised settings, to generate nodes? content embeddings.
%%Furthermore, we employed convolutional layers in a semi-supervised manner to overcome the scarcity of training data.
%In the end, we conducted empirical evaluation on several real-world datasets and showed that our model improved the accuracy of GNNs.

In this paper, 
we proposed novel methods that augment nodes' embeddings with their content information at higher GNN layers.
%we suggested augmenting nodes' embeddings by embeddings generating from their content in different GNN layers.
In our methods, for each node
a {\em structural} embedding and a {\em content} embedding are computed and combined using a combination layer,
to form the embedding of the node at a given layer.
%using a GNN and a {\em content} embedding are computed for each node.
%These two are combined using a combination layer to form the embedding of a node at a given layer. 
We presented
techniques such as using an auto-encoder or building a content graph to generate nodes' {\em content} embeddings.
%an embedding is extracted from the content of each node and concatenated with the embedding generated from the graph neural network. An unsupervised dimension reduction method based on auto-encoders is employed to reduce the dimensions of the generated embeddings to a desired value.
%Our method is independent of the specific GNN used and can be aligned with any of them.
We conducted experiments over several real-world datasets and showed that our models considerably improve the accuracy of GNN models.
In particular, they show high performance when graph nodes have a high number of neighbors, where initial node features are forgotten during the message-passing process.
% Even though our models gain improvements in sparse datasets, they demonstrate significant performance when the average degree increases.

\bibliographystyle{plain}
\bibliography{allpapers.bib}% common bib file

\end{document}